\documentclass[lettersize,journal]{IEEEtran}
\usepackage{amsmath,amsfonts}
\usepackage{algorithmic}
\usepackage{array}
\usepackage[caption=false,font=normalsize,labelfont=sf,textfont=sf]{subfig}
\usepackage{textcomp}
\usepackage{stfloats}
\usepackage{url}
\usepackage{verbatim}
\usepackage{graphicx}
\def\BibTeX{{\rm B\kern-.05em{\sc i\kern-.025em b}\kern-.08em
    T\kern-.1667em\lower.7ex\hbox{E}\kern-.125emX}}
\usepackage{balance}

\usepackage{cite}
\usepackage{amsfonts}
\usepackage{amsmath}
\usepackage{booktabs}
\usepackage{color}
\usepackage{url}
\usepackage{graphicx}
\usepackage{bbding}
\usepackage{multirow}
\usepackage{makecell}
\usepackage{newfloat}
\usepackage{listings}
\usepackage[table,xcdraw]{xcolor}
\usepackage[utf8]{inputenc}
\usepackage[T1]{fontenc}
\usepackage[marginal]{footmisc}
\usepackage{amssymb}

\begin{document}
\title{Pixel Difference Convolutional Network for RGB-D Semantic Segmentation}
\author{$\text{Jun Yang}^{\dagger}$, $\text{Lizhi Bai}^{\dagger}$, $\text{Yaoru Sun}^{*}$, $\text{Chunqi Tian}^{*}$, Maoyu Mao, Guorun Wang

\thanks{Jun Yang, Lizhi Bai, Yaoru Sun, Chunqi Tian, Maoyu Mao, Guorun Wang are with Department
    of Computer Science and Technology, Tongji University, Shanghai, China
    (e-mail: \{junyang, bailizhi, yaoru, maomy, 1950575\}@tongji.edu.cn, tianchunqi@163.com).
    Jun Yang is also with the Laboratoire des Systèmes Perceptifs (UMR8248),
École normale supérieure,
PSL Research University, Paris, France.}
    }


\maketitle
\footnote{
{$*$} Corresponding author. \\
{$\dagger$} Equal contribution.}

\maketitle

\begin{abstract}
RGB-D semantic segmentation can be advanced with convolutional neural networks due to the availability of Depth data. Although objects cannot be easily discriminated by just the 2D appearance, with the local pixel difference and geometric patterns in Depth, they can be well separated in some cases. Considering the fixed grid kernel structure, CNNs are limited to lack the ability to capture detailed, fine-grained information and thus cannot achieve accurate pixel-level semantic segmentation. To solve this problem, we propose a Pixel Difference Convolutional Network (PDCNet) to capture detailed intrinsic patterns by aggregating both intensity and gradient information in the local range for Depth data and global range for RGB data, respectively. Precisely, PDCNet consists of a Depth branch and an RGB branch. For the Depth branch, we propose a Pixel Difference Convolution (PDC) to consider local and detailed geometric information in Depth data via aggregating both intensity and gradient information. For the RGB branch, we contribute a lightweight Cascade Large Kernel (CLK) to extend PDC, namely CPDC, to enjoy global contexts for RGB data and further boost performance. Consequently, both modal data's local and global pixel differences are seamlessly incorporated into PDCNet during the information propagation process. Experiments on two challenging benchmark datasets, $i.e.$, NYUDv2 and SUN RGB-D reveal that our PDCNet achieves state-of-the-art performance for the semantic segmentation task.
\end{abstract}


\begin{figure}
    \centering
    \includegraphics[scale=0.25]{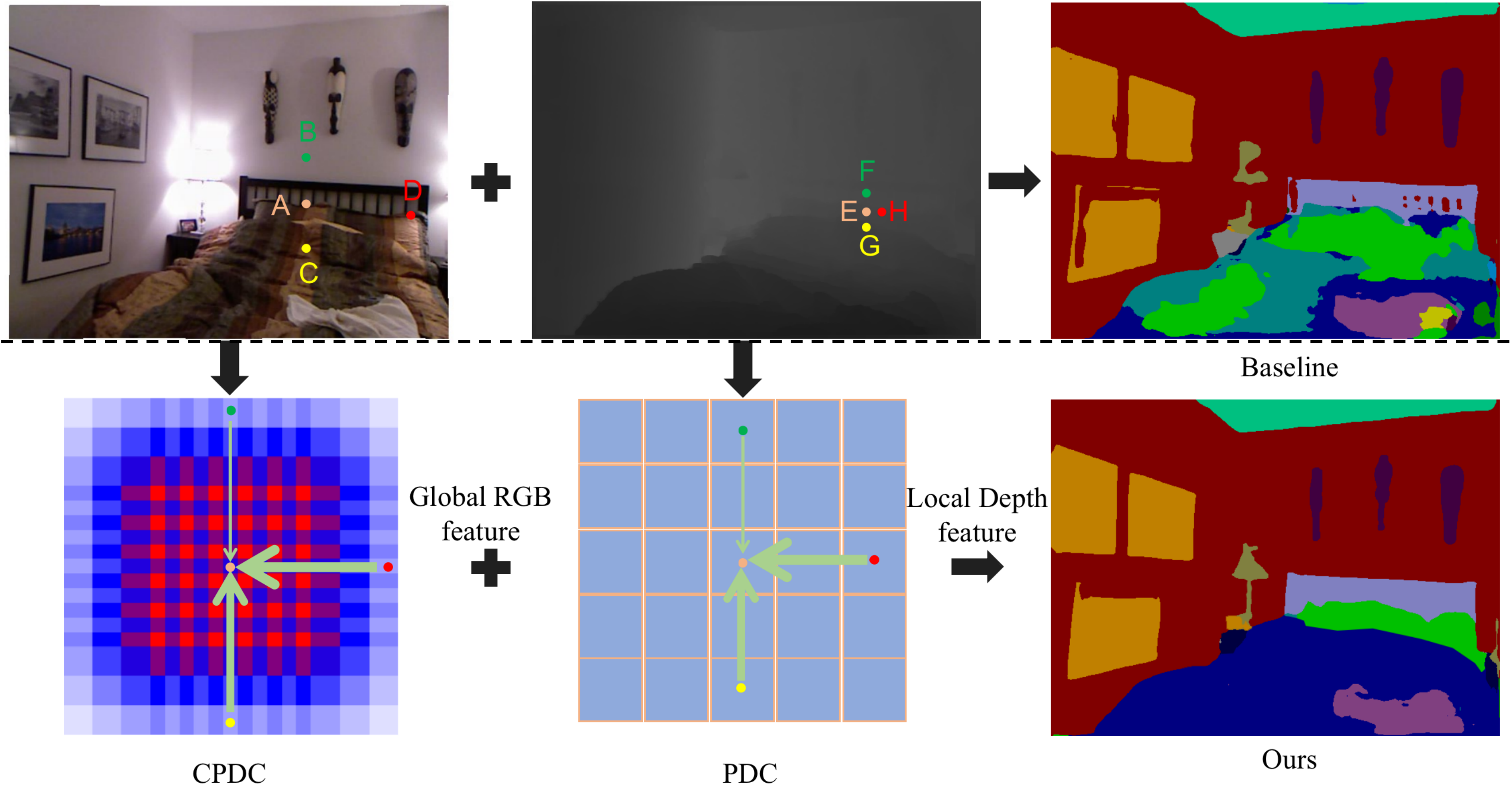}
    \caption{Illustration of our PDCNet. In the RGB image, pixels A and D are the edges of pillow; B and C belong to other categories. RGB images reflect overall spatial information. In the Depth image, G is the junction of the bed and the pillow. While the pillow and bed have similar visual characteristics in the RGB image, they can be distinguished in Depth. Based on this, we propose PDC to capture local subtle pixel differences in RGB and CPDC to focus on long-range pixel dependence for RGB.}
    \label{hard_sample2}
\end{figure}

\section{Introduction}
Semantic segmentation infers semantic labels of every pixel in a scene. With the help of 3D sensors, RGB-D data boosts the advancement of RGB-D semantic segmentation. Specifically, RGB image captures photometric appearance properties in projected image space, while Depth map provides local geometry appearance cues that are complementary to RGB images \cite{huang2022adaptive, liu2022geometrymotion}.

However, the convolution operator that is widely adopted for consuming RGB data might be suboptimal for processing the Depth data. For example, \textbf{pillows on a bed with similar color of bed in Fig.\ref{hard_sample2}.} It is evident that it is not feasible to distinguish pillows based solely on 2-dimensional features such as colors and shapes. 
In contrast with RGB images, without being limited by the similar indiscernible appearance, Depth maps have a foreground object that is distinct from the background, which theoretically leads to better segmentation performance, compared to RGB images only. In fact, Depth contours are cleaner than RGB contours and tend to better describe the edge information of objects for better separation of objects \cite{zhao2022self}. In contrast, RGB images are frequently utilized for providing global information \cite{wang2018non}. As a result, it is vital to take advantage of the two complementary modalities to improve semantic segmentation performance.

Recently, researchers make remarkable efforts on RGB-D semantic segmentation tasks \cite{he2017std2p,jiang2018rednet}. They conduct fusion models which leverage RGB images and Depth images for the better understanding of the two complementary modalities. In some research, Depth maps are applied as an extra channel for the RGB images in the early stage of the model \cite{he2017std2p,husain2016combining}. Similarly, the fusion method can also be designed by fusing the features extracted from the RGB and Depth data \cite{jiang2018rednet, zhu2022vpfnet}. However, these fusion methods either lack the non-linear feature fusion ability or treat both modal images equally by engaging the same contribution from both modal data, regardless of the intrinsic distribution gap or information loss, which is obviously not appropriate.

Furthermore, due to the fixed grid kernel structure, convolutional networks are limited by their lack of ability to encode detailed gradient information and therefore cannot focus on pixel difference so as to accurately achieve pixel-level segmentation. Incorporating the geometric information from Depth images into CNN is essential yet challenging. 

To mitigate the problems mentioned above, we propose a Pixel Difference Convolutional Network (PDCNet) for RGB-D segmentation. Two mechanisms, namely Pixel
Difference Convolution (PDC) and Cascade Large Kernel (CLK)-based PDC (CPDC), are introduced to improve the ability of two modal data interactions in this task.

The PDC augments the standard convolution with a pixel difference term while aggregating local intensity and gradient information of objects in Depth images. According to the principle of PDC, this pixel difference operation can also solve the problem of noise in the depth map.
The CLK, consisting of three small convolutions to obtain a field of perception of $21 \times 21$, has a significantly lower computational cost than the corresponding convolution of size 21.
CLK not only involves the advantages of long-range dependence in 2D structure, but also achieves channel-wise adaptability. After extending PDC with CLK, the CPDC combines the advantage of capturing local structure information of PDC and long-range dependence of CLK. The CPDC is employed for RGB data to enjoy global contextual information. As a result, the PDC and CPDC exploit the characteristics of 
the two modal data and complement each other effectively. In summary, the contributions of this paper are fourfold as below: 

$\bullet$ We propose a PDC for considering the subtle intensity and gradient information in local regions for Depth data, which integrates the pixel difference operators into the convolutional operations. The principle of PDC also shows its ability to overcome the noise of the depth map.

$\bullet$ We propose a CLK to achieve long-range receptive fields. The combination of CLK and PDC, namely CPDC, combines both advantages to extend PDC to enjoy the global information of RGB images.

$\bullet$ Both PDC and CPDC are seamlessly plugged and
play in existing CNNs to form a two-branch PDCNet. PDCNet takes advantage of the characteristics of the two modal data, generating a more robust modeling capacity.

$\bullet$ Our proposed model achieves a new state-of-art performance on NYUDv2 and SUN RGB-D datasets.

\section{Related Work}

\subsection{Context information Aggregation}
Contextual information is commonly utilized in semantic segmentation networks to enhance feature representation. Non-local network \cite{wang2018non} focuses on spatial information distribution, which guarantees that a pixel at any position perceives
contextual information from all pixels. Atrous spatial
pyramid pooling (ASPP) is proposed in Deeplabv2 \cite{chen2017deeplab} to enhance contextual information utilizing different dilation convolutions. Further, DenseASPP \cite{yang2018denseaspp} brings the dense connections into ASPP to encode the multi-scale features. CCNet \cite{huang2019ccnet} captures full-image contextual information when introducing less computation and memory cost. PSA \cite{zhao2018psanet} builds an attention map that aggregates contextual information for individual points in a specific and adaptive manner. Nie \textit{et~al.} \cite{nie2022mign} propose an encoder-decoder architecture and a MRIF module to generate high-resolution segmentation results by considering both details and boundary information. Gao \textit{et~al.} \cite{gao2022fbsnet} propose a bottleneck residual unit where the convolutions have different dilation rates to achieve rich contextual information.

Our PDCNet captures contextual information by aggregating both intensity and gradient subtle pixel-wise difference/similarity in local range for Depth data and global range for RGB data, respectively.

\subsection{RGB-D Semantic Segmentation} 
Recent advances in semantic segmentation benefit a lot from combining such two complementary \cite{ren2012rgb,silberman2012indoor,jiao2019geometry,cao2021shapeconv,gupta2013perceptual,khan2016integrating}. Many works simply concatenate the features of RGB and Depth images to enhance the semantic information of each pixel \cite{silberman2012indoor,ren2012rgb}. Typically, the RGB-D fusion methods are classified into early, middle, and late fusion. As an early stage fusion method, Cao \textit{et~al.} \cite{cao2021shapeconv} firstly concatenate the RGB and Depth data and then decompose them into a shape and a base component for extracting semantic information by individual corresponding learnable weights. However, two types of data contain inconsistent features and cannot be processed by shared network feature extractors. Jiao \textit{et~al.} \cite{jiao2019geometry} apply the encoder-decoder architecture to make
full use of the both 2D appearance and 3D geometry information in late stage by only taking one single RGB image
as input. In this method, the interaction between both modal data is insufficient and the learned embeddings are gradually compressed and even lost. The middle stage fusion strategy outperforms the aforementioned methods, which overcomes the drawback of early stage and late stage fusion strategies by interacting the intermediate information of the two different modalities. Chen \textit{et~al.} \cite{chen2020bi} present a cross-modality guided encoder to fuse each pair of
feature maps by the SA-Gate and propagate to the next stage for further feature transformation. This method reduces the distribution gap in the middle stage and achieves ample interaction from multi-modal features. Chen \textit{et~al.} \cite{chen2021spatial} a spatial information-guided convolution is proposed to enhance the spatial adaptability, which incorporates geometric information into the feature learning
process by generating spatially adaptive convolutional weights. Zhang \textit{et~al.} \cite{zhang2022tube} propose a coarse-to-fine decoder to generate the feature based on the transformer mechanism, which achieves pixel-level
multi-task prediction.

Our PDCNet applies a two-branch structure, where each branch focuses on extracting modality-specific features. Specifically, the Depth branch focuses on local geometric and illumination-independent features from Depth images by PDC module. RGB branch extracts global color and texture information by CPDC module. Attentive RGB and contextual features generated by the CPDC and PDC are fused at each resolution stage in the encoders.

\section{Methodology}
In this section, we first present the formulation of our proposed \textbf{P}ixel \textbf{D}ifference \textbf{C}onvolution (PDC), then present the \textbf{C}ascade \textbf{L}arge \textbf{K}ernel (CLK) and the combined CPDC to implicitly improve the semantic segmentation performance. At last, we present the \textbf{P}ixel \textbf{D}ifference \textbf{C}onvolutional \textbf{N}etwork (PDCNet) for our task.

\subsection{Overview}

The discrepancy of adjacent pixels is usually subtle and occurs in local regions, which is not easy to be captured by vanilla convolutional networks. To fill this gap, we propose a pixel difference convolution to capture subtle pixel difference, which is a straightforward solution for refined RGB-D semantic segmentation.

The existing methods either treat RGB and Depth images equally or mainly consider the Depth data. We believe that the depth map is more responsive to object boundary information due to the geometry characteristics, while RGB better reflects the global image information. In our paper, the PDC is applied to capture local information such as edges and boundaries for Depth data. On the contrast, we propose a CLK embedded into our PDC, namely CPDC, to achieve a long-range pixel-wise relationship for RGB images. Both PDC and CPDC apply $depthwise$ $separable$ $convolution$ \cite{chollet2017xception} structure to reduce computation and memory.


\begin{figure*} 
    \centering
    \includegraphics[scale=0.65]{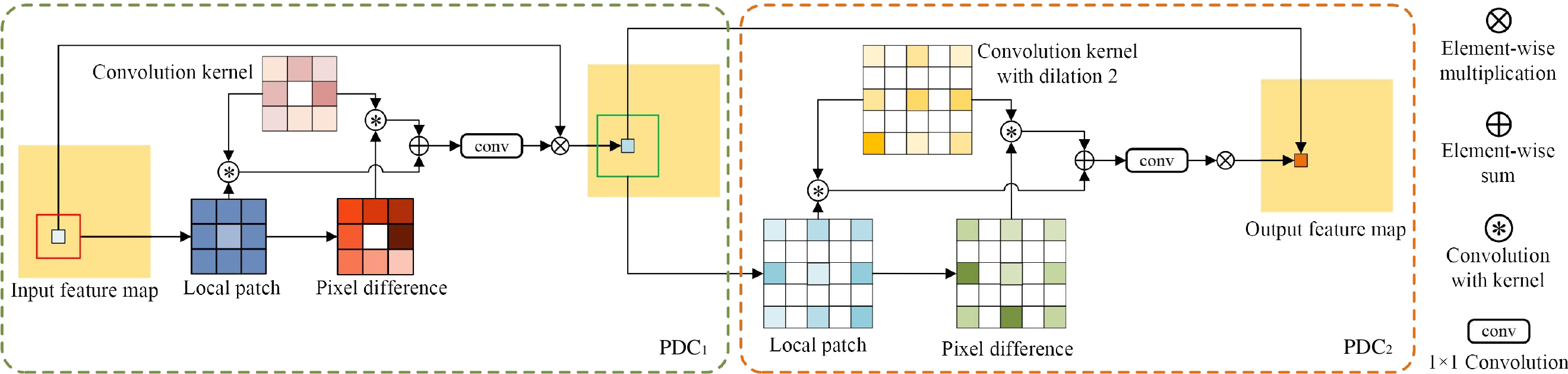}
    \caption{The structure of PDC$_1$ (left) and PDC$_2$ (right). Without loss of
generality, we show the $3\times 3$ convolution kernel for the PDC$_1$ and the $3\times 3$ convolution kernel with dilation of 2 for the PDC$_2$. The operation combining PDC$_1$ and PDC$_2$ contributes the CPDC.}
    \label{combination}
\end{figure*}

\subsection{Pixel Difference Convolution}
\label{sec-DDCA}
\begin{figure}
    \centering
    \includegraphics{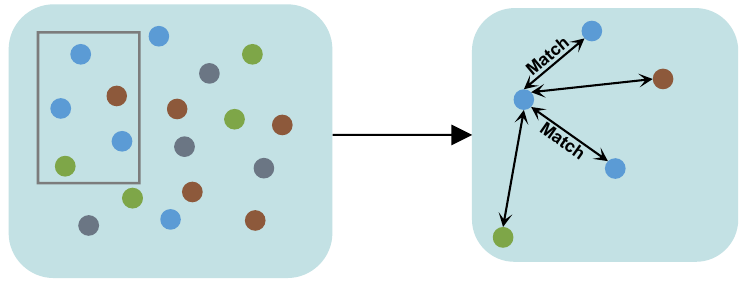}
    \caption{In pixel-difference calculations, intra-class compactness and inter-class dispersion are motivated by pulling closer pixels within a class and pushing away pixels within another class.}
    \label{difference}
\end{figure}

The sense should be emphasized that the pixels with the same semantic labels have more substantial pixel-wise similarity \cite{zhang2020supervised}. As shown in Fig.~\ref{difference}, the pixel-wise difference can be applied to force pixels with more consistent
geometry to make more contributions to the corresponding output. This relationship between different points is obtained by our PDC.

\subsubsection{Vanilla Convolution}
An ordinary 2D convolution computes the weighted sum of the local grid. There
are two main steps in the 2D convolution: 1) sampling local receptive ﬁeld region $\mathbb{R}$ over the input feature map $\mathbf{x}$;
2) aggregating sampled values through weighted summation.
For each pixel location $\mathbf{p}_0$, the output feature map $\mathbf{y}$ is:

\begin{equation}
    \mathbf{y}(\mathbf{p}_0) = Conv(\mathbf{w},\mathbf{p}_n) = \sum_{\mathbf{p}_n \in \mathbb{R}} \mathbf{w}(\mathbf{p}_n)\cdot \mathbf{x}(\mathbf{p}_0 + \mathbf{p}_n),
    \label{vanilla}
\end{equation}
where $\mathbf{w}$ is the convolution kernel, $\mathbf{p}_0$ indicates current location on both input and output feature maps while $\mathbf{p}_n$ enumerates the locations in $\mathbb{R}$.

\subsubsection{Vanilla Convolution Meets Pixel Difference}
To exploit the correlation between pixels, we introduce pixel difference into vanilla convolution to enhance its representation and generalization capacity. The new convolution formulation also consists of two main steps, where the aggregation step is different from that in vanilla convolution. As illustrated in the left part of Fig.~\ref{combination}, the pixel difference convolution operation, when applied, focuses on aggregating the gradient information from surrounding pixel points of a given pixel point. We directly apply subtraction operation to find original pixel difference, Eq.(\ref{vanilla}) henceforth becomes:
\begin{equation}
    \begin{split}
        \mathbf{y}(\mathbf{p}_0) &= Conv(\mathbf{w},\Delta \mathbf{p}) \\
        &= \sum_{\mathbf{p}_n \in \mathbb{R}} \mathbf{w}(\mathbf{p}_n)\cdot \left(\mathbf{x}(\mathbf{p}_0 + \mathbf{p}_n) - \mathbf{x}(\mathbf{p}_0)\right).
    \end{split}
    \label{pixel_difference}
\end{equation}
Eq.(\ref{pixel_difference}) supports that the pixels with the same semantic labels have more substantial pixel-wise similarity. We make use of this similarity to force pixels with more consistent geometry to make more contributions to the corresponding output. 

\subsubsection{Details of Pixel difference convolution}
It is obvious that both fine-grained gradient and intensity semantic information are vital for RGB-D semantic segmentation task. Based on this, we aim to combine the vanilla convolution with pixel difference convolution to provide a more robust feature capture capacity. For this purpose, we generalize the pixel difference convolution as: 

\begin{equation}
    \begin{split}
    \mathbf{y}(\mathbf{p}_0) &= \alpha \cdot Conv(\mathbf{w},\Delta \mathbf{p}) + (1-\alpha) \cdot Conv(\mathbf{w}, \mathbf{p}_n)\\
    &= \alpha \cdot \underbrace{\sum_{\mathbf{p}_n \in \mathbb{R}} \mathbf{w}(\mathbf{p}_n)\cdot \left(\mathbf{x}(\mathbf{p}_0 + \mathbf{p}_n) - \mathbf{x}(\mathbf{p}_0)\right)}_{\text{pixel difference term}} \\
    &+(1-\alpha) \cdot \underbrace{\sum_{\mathbf{p}_n \in \mathbb{R}} \mathbf{w}(\mathbf{p}_n)\cdot \mathbf{x}(\mathbf{p}_0 + \mathbf{p}_n)}_{\text{vanilla term}}.
    \label{pdc}
    \end{split}
\end{equation}

In this context, $\alpha$ is used to adjust the contribution of fine-grained gradient-level changes and intensity-level changes for semantic information. The higher the $\alpha$ value, the more important of pixel differences are, and vice versa. In this paper, we set $\alpha$ as the learnable parameter. Additionally, we also set the $\alpha$ as default configurations, $i.e.$, varying from 0 to 1, to show the effectiveness of pixel difference term. The impact of $\alpha$ is further explained in the ablation studies.

We name the formulation of \textbf{P}ixel \textbf{D}ifference \textbf{C}onvolution (PDC) displayed in Eq.(\ref{pdc}) as \textbf{PDC} for short. 
We choose the kernel size of $5 \times 5$ with dilation of 1 for PDC to capture local fine-grained information.
\subsubsection{Implementation for PDC} To better implement PDC, we follow the form in Pytorch \cite{yu2020searching}.
After rewriting the form of Eq.(\ref{pdc}) through simple compose and merge operation, the PDC can be easily implemented in CNNs. Eq.(\ref{pdc}) becomes:

\begin{equation}
    \begin{split}
    \mathbf{y}(\mathbf{p}_0) = &-\alpha \cdot \underbrace{\sum_{\mathbf{p}_n \in \mathbb{R}} \mathbf{w}(\mathbf{p}_n) \cdot \mathbf{x}(\mathbf{p}_0)}_{\text{pixel difference term}} \\
    &+\underbrace{\sum_{\mathbf{p}_n \in \mathbb{R}} \mathbf{w}(\mathbf{p}_n)\cdot \mathbf{x}(\mathbf{p}_0 + \mathbf{p}_n)}_{\text{vanilla term}}.
    \label{pdc_new}
    \end{split}
\end{equation}
It should be noted that $\mathbf{w}(\mathbf{p}_n)$ is shared between the pixel difference term and vanilla convolution term. Compared with Eq.(\ref{pdc}), Eq.(\ref{pdc_new}) can be easily performed. Subsequently, given an input feature map $\mathbf{F} \in \mathbb{R}^{H \times W \times C}$, the output utilizing PDC is generated as: 

\begin{equation}
    \mathcal{O}_\text{pdc}=
    Conv_{1\times 1}\big( \text{PDC} (\mathbf{F}) \big) \otimes \mathbf{F},
\label{pdc_out}
\end{equation}
where $Conv_{1\times1}$ is a channel convolution for achieving the channel-wise adaptability.

\begin{figure}[t]
    \centering
    \begin{minipage}[c]{0.32\linewidth}
        \centering
        \centerline{
        \includegraphics[width=1.3in]{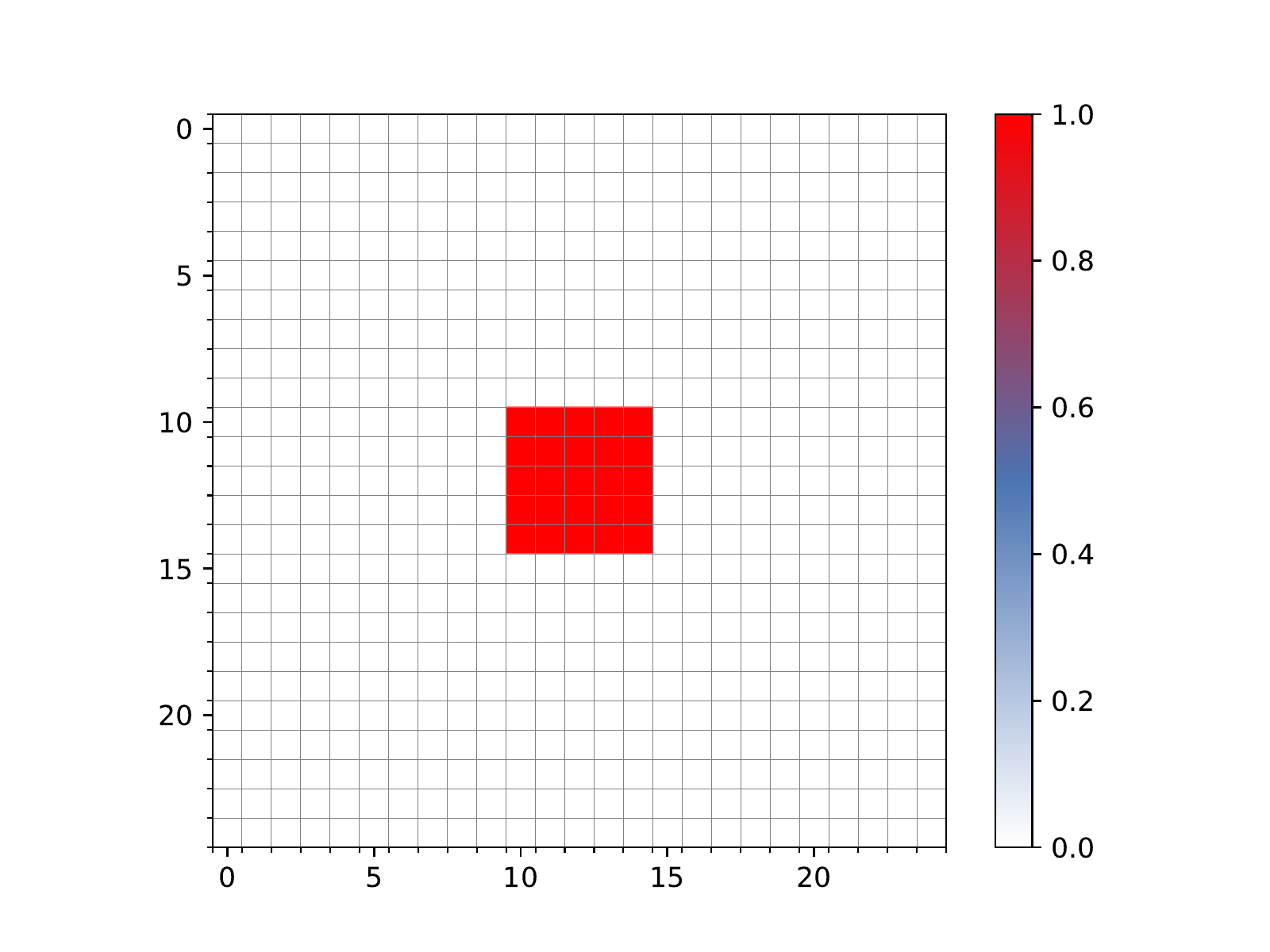}
        }
        \centerline{(a)}
    \end{minipage}
    \begin{minipage}[c]{0.32\linewidth}
        \centering
        \centerline{
        \includegraphics[width=1.3in]{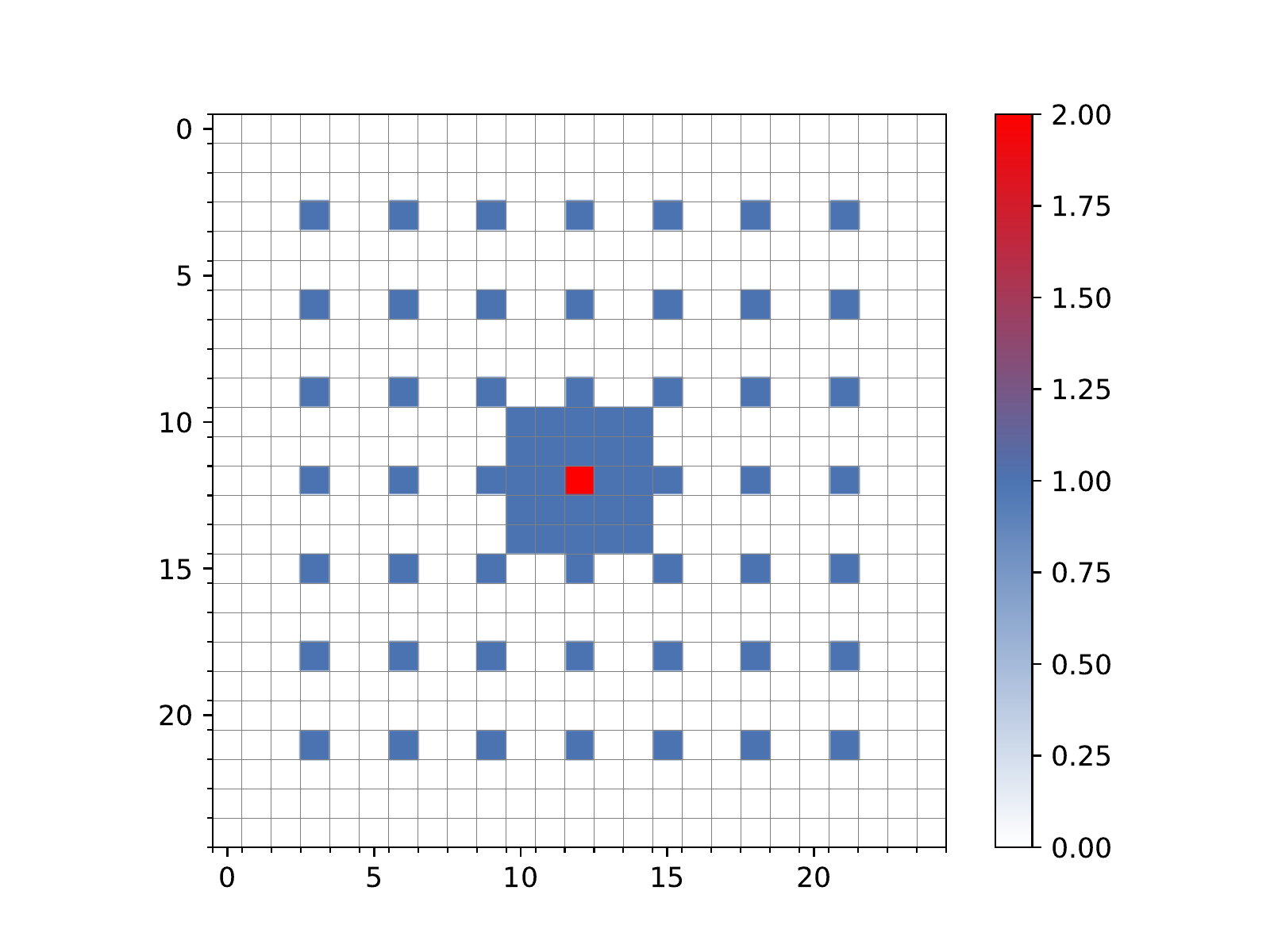}
        }
        \centerline{(b)}
    \end{minipage}
    \begin{minipage}[c]{0.32\linewidth}
        \centering
        \centerline{
        \includegraphics[width=1.3in]{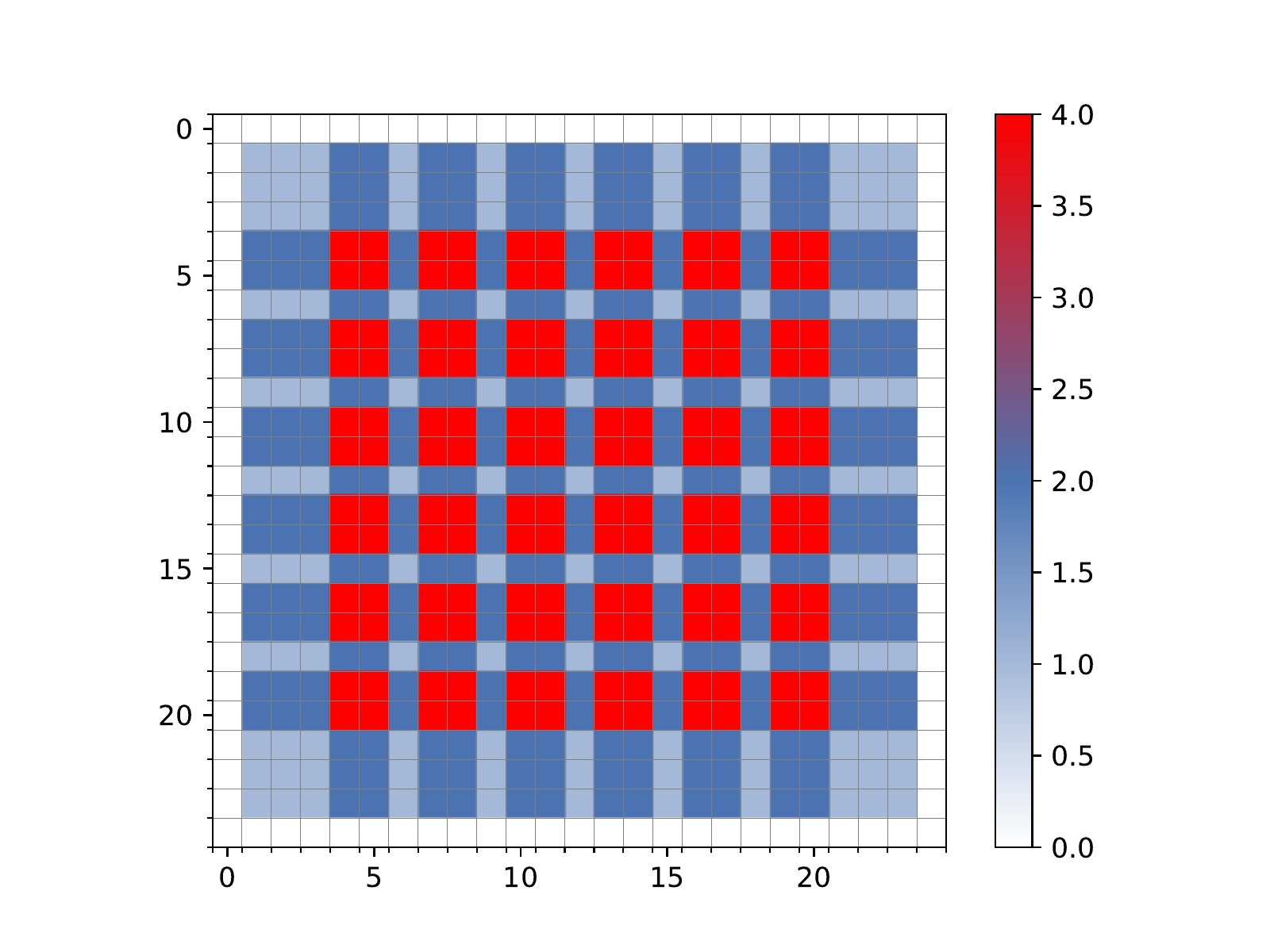}
        }
        \centerline{(c)}
    \end{minipage}

     \caption{Illustration of parallel mode and cascade mode. (a) indicates $Conv_{5\times 5}$, (b) indicates parallel mode: $Conv_{5\times 5}(\cdot) +Conv_{7\times 7}(\cdot)$ and (c) indicates cascade mode: $Conv_{7\times 7}(Conv_{5\times 5}(\cdot)$). The color bar indicates the number of times each pixel point in the feature map is used.}
\label{ELKA}
\end{figure}

\subsection{Cascade Large Kernel} 
As we mentioned before, different from Depth maps, RGB images are better suited to provide contextual information and spatial adaptability within the global field of view. To this purpose, we propose a CLK to provide long-range receptive fields.
We do not utilize the CLK directly; instead, the proposed CLK is the key component applied to extend the PDC for RGB data.  

\subsubsection{Implementation for CLK}
Our CLK consists of three convolutions, that is, \textit{a depth-wise local convolution} ($Conv_{5\times 5}$, dilation of 1), \textit{a depth-wise long-range convolution} ($Conv_{7\times 7}$, dilation of 3, and \textit{a point-wise convolution} ($1\times1$ convolution) to approximate a $21 \times 21$ convolution. Under this setting, CLK can effectively achieve both local information as well as long-range dependence. Notably, the dilation of $7 \times 7$ convolution can be filled by the $5\times 5$ convolution,
which means the CLK does not cause any information loss compared to other long-ranged feature extraction methods using ordinary dilated convolution, such as PPM \cite{zhao2017pyramid} and ASPP \cite{chen2017deeplab}.

    


\subsubsection{Cascade mode vs. Parallel mode} To prove the effectiveness of CLK, we compare two common modes, that is, parallel mode and cascade mode. Since the channel convolution do not change the receptive field, we just show the receptive field of the two modes in Fig.~\ref{ELKA}. 
The output feature maps generated by the two modes can be written as:
\begin{equation}
    \begin{split}
    & \mathcal{O}_\text{parallel} \rightarrow Conv_{{5\times 5}}(\mathbf{F})+Conv_{7\times 7}(\mathbf{F}),\\
    & \mathcal{O}_\text{cascade} \rightarrow Conv_{7\times 7}\left(Conv_{5\times 5}(\mathbf{F})\right). 
    \end{split}
    \label{mode}
\end{equation}

Here, $\mathbf{F} \in \mathbb{R}^{H \times W \times C}$ is an input feature map. Obviously, compared with the receptive field of parallel mode, cascade mode has a large receptive field. Additionally, parallel mode loses a large portion of local information due to the dilation. This proves that the cascade mode we apply is better. We name the cascade mode as \textbf{C}ascade \textbf{L}arge \textbf{K}ernel (CLK).

\subsubsection{CLK vs. large kernels}
Our CLK differs from ordinary large kernel convolution in two ways. Firstly, the CLK achieves the receptive field of $21 \times 21$ utilizing a $5 \times 5$ convolution, a $7 \times 7$ convolution and a $1 \times 1$ convolution, the computational cost significantly less than the large kernel with size of 21. Notably, the dilation of $7 \times 7$ convolution can be filled by the $5 \times 5$ convolution, which means the CLK do not lacks any information like ordinary dilated convolution. Secondly, the CLK also achieves channel-wise adaptability shown in Eq.~\ref{pdc_out} compared with an ordinary large kernel. 
The proposed CLK is further employed to extend PDC to build CPDC for RGB data.

\subsection{CPDC: Extend PDC with CLK.}
Since PDC focuses on both fine-grained gradient-level and intensity-level semantic information in local regions from Depth data, we further apply CLK to extend PDC to enjoy global information for RGB images. 

From Eq.(\ref{pdc_new}), the field size of convolution is input dependent, we hence directly replace it with our CLK. The combination of PDC and CLK, $i.e.,$ CPDC is demonstrated in Fig.~\ref{combination}, and the formulation of CPDC can be summarized as: 
\begin{equation}
\mathbf{F}_{out} = \text{PDC}_{7\times 7}\left(\text{PDC}_{5\times 5}(\mathbf{F})\right).
\end{equation}
Notably, the PDC formulation in Eq.(\ref{pdc}) applies the a convolution with size of $5\times 5$ with dilation of 1. Here, $\text{PDC}_{5 \times 5}$ denotes the PDC operation with $Conv_{5\times 5}$: $5\times 5$ convolution, dilation of 1 and PDC$_{7\times 7}$ denotes the PDC operation with $Conv_{7\times 7}$: $7\times 7$ convolution, dilation of 3, respectively.

Similar to Eq.(\ref{pdc_out}), the corresponding output is formulated as:
\begin{equation}
    \mathcal{O}_\text{cpdc} =
    Conv_{1\times1}(\mathbf{F}_{out})\otimes \mathbf{F},
    \label{cpdc_fun}
\end{equation}

where $Conv_{1\times1}$ is a channel convolution for achieving the channel-wise adaptability. When embedded into the PDC, CPDC enjoys the advantages of both fine-grained gradient-level and intensity-level semantic information in a satisfied receptive field.

\begin{figure*}[t]
    \centering
        \centering
        \centerline{
        \includegraphics[scale=0.5]{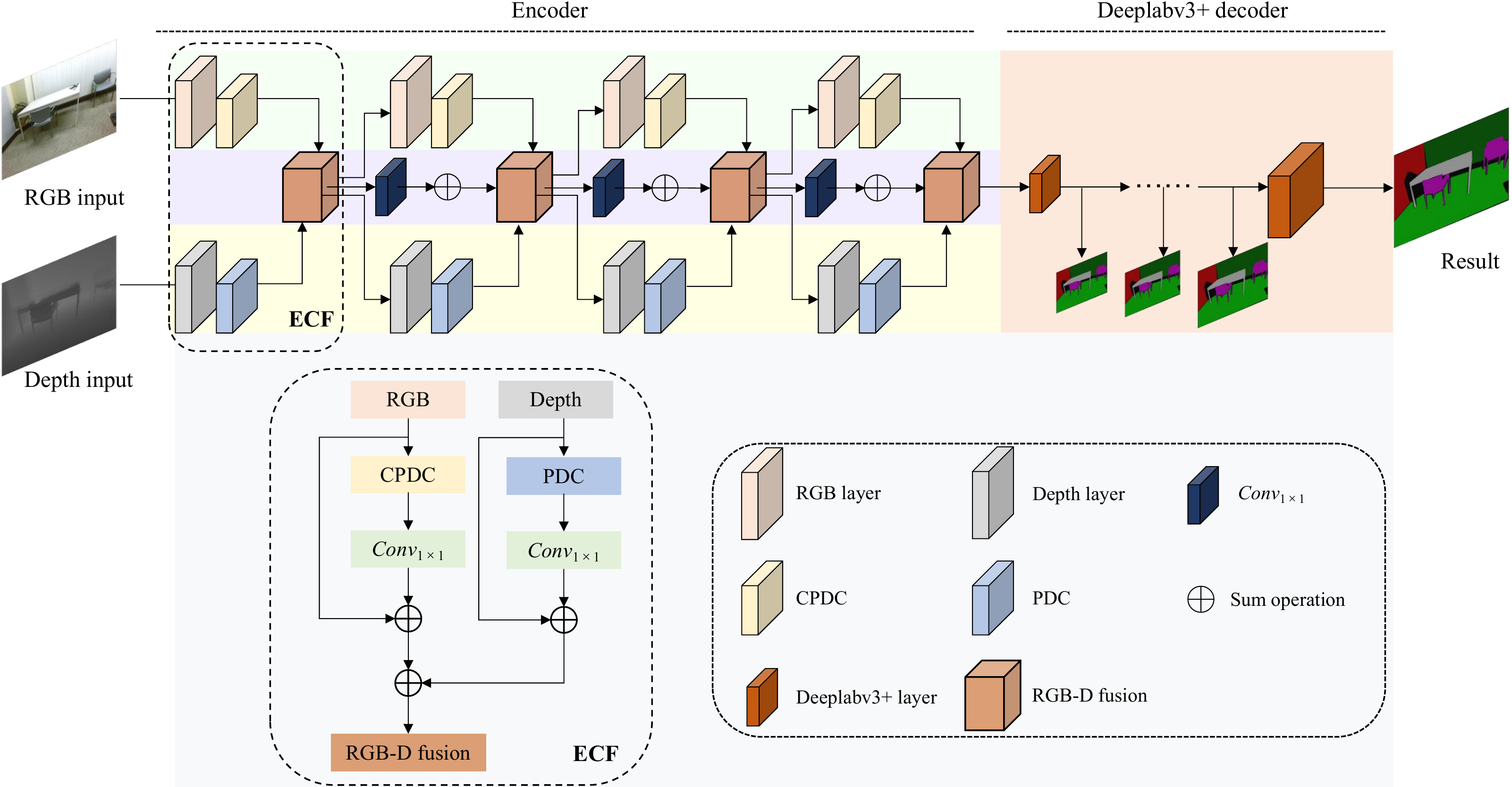}
        }
     \caption{The overview of PDCNet. The network is a two-branch structure, which consists of two ResNet-101 encoders. The CPDC and PDC are equipped after each block of ResNet-101 encoder in the RGB branch and Depth branch, respectively. During training, output feature maps of RGB layer and Depth layer are fused through a ECF module and then fed into the next stage.}
\label{model_architecture}
\end{figure*}

\subsection{PDCNet Architecture}
\subsubsection{The Overall Architecture}
We adopt the DeepLabv3+ \cite{chen2018encoder} to train our baseline model, where the encoder is ResNet-101 \cite{he2016deep} and keep its original DeepLabv3+ settings. We apply a two-branch structure in our PDCNet, one for RGB (green part in Fig.~\ref{model_architecture}) and another for Depth data (yellow part in Fig.~\ref{model_architecture}). Instead of treating RGB and Depth data equally, we apply the PDC for the Depth branch and CPDC for the RGB branch. The proposed PDC and CPDC are generalized approaches that are easily plugged into each block of backbone layers in the two different branches. The feature representations of both branches are fused in each stage of the backbone network. In addition, we also employ multi-scale feature fusion strategy, where low-level and high-level semantic features are combined to produce rich maps (purple part in Fig.~\ref{model_architecture}). The architecture of our PDCNet is shown in Fig.~\ref{model_architecture}.

\subsubsection{Efficient Cross-modal Fusion (ECF)}
Firstly, the PDC is employed for Depth data in each ResNet-101 block in Depth branch, while CPDC is employed for RGB data in RGB branch. Secondly, for cross-modal fusion, the complementary geometric information generated by PDC is fused into the RGB encoder at each stage of the ResNet-101. 

Assume that the RGB and Depth features of $l$-th layer are $\mathbf{F}_\text{rgb}^{(l)} \in \mathbb{R}^{H\times W \times C}$ and $\mathbf{F}_\text{Depth}^{(l)} \in \mathbb{R}^{H\times W \times C}$, respectively. 
The contribution on RGB and
Depth data as follows:

\begin{equation}
    \hat{\mathbf{F}}_\text{Depth}^{(l)}= \text{PDC}\left(\mathbf{F}_\text{Depth}^{(l)} \right), 
    \hat{\mathbf{F}}_\text{rgb}^{(l)} = \text{CPDC}\left(\mathbf{F}_\text{rgb}^{(l)} \right).
\label{contribution}
\end{equation}
We set learnable parameters $\eta$ and $\lambda$ for the better RGB and Depth features fusion. The two parameters  $\eta$ and $\lambda$ are not directly related as it is not reasonable to directly control the fusion ratio of the two modal data. Consequently, the RGB-D fusion map of $l$-th layer are generated as:
\begin{equation}
  \begin{split}
    \mathbf{F}_\text{fusion}^{(l)} &=\eta \left( Conv_{1\times1}(\hat{\mathbf{F}}_\text{rgb}^{(l)})\otimes \hat{\mathbf{F}}_\text{rgb}^{(l)} \oplus \mathbf{F}_\text{rgb}^{(l)}) \right)\\ &+ \lambda \left( Conv_{1\times1}(\hat{\mathbf{F}}_\text{Depth}^{(l)})\otimes \hat{\mathbf{F}}_\text{Depth}^{(l)} \oplus \mathbf{F}_\text{Depth}^{(l)}) \right ),
  \end{split}
\end{equation}
where $Conv_{1\times1}$ is a channel convolution for achieving the channel-wise adaptability.

\subsubsection{Loss Function}
Our proposed model is trained on the training test 
\begin{equation}
    \begin{split}
    \{ (\mathbf{X}_{rgb}^i, \mathbf{X}_{Depth}^i, \mathbf{Y}^i)|\mathbf{X}_{rgb}&\in \mathbb{R}^{H\times W\times 3},  \mathbf{X}_{Depth}\in \mathbb{R}^{H\times W},\\
    &\mathbf{Y}^i \in L^{H\times W}, i=\{1,\cdots, M\} \},
    \end{split}
\end{equation}
where $\mathbf{Y}$ represents the label and $M$ represents the number of classes.

The loss function is cross-entropy, which is defined as:

\begin{align}
    \mathcal{L} &= \frac{-1}{N}\sum_i^N\ell(yi,\hat{y}i),\\
    \ell(yi,\hat{y}i) &= -\sum_i^M y_n log(p_n),
\end{align}
where $y_i$ and $\hat{y}i$ indicates the true label and the predictive label of samples. 
$N$ denotes the pixels in the training data and $p_n$ denotes the possibility when the current pixel belongs to class n.

\begin{table*}[]
\centering
\caption{Performance comparison with the state-of-the-art methods on NYUDv2 dataset. IS: Information source. `*': multi-scale strategy.}
\resizebox{0.95\linewidth}{!}{
\begin{tabular}{c|c|c|c|c|c|c|c|c|c}
\toprule
Method                      & Backbone  & IS.   & Pixel Acc. & mIoU & Method                            & Backbone    & IS.   & Pixel Acc. & mIoU                           \\ \midrule
FCN \cite{long2015fully}           & VGG16     & HHA   & 65.4       & 34.0 & RDF \cite{park2017rdfnet}*               & ResNet152   & HHA   & 76.0       & 50.1                           \\ 
LSD-GF \cite{cheng2017locality}    & VGG16     & HHA   & 71.9       & 45.9 & DeepLab-LFOV \cite{chen2017deeplab}*     & VGG16       & Depth & 70.3       & 39.4                           \\ 
D-CNN \cite{wang2018}         & VGG16     & Depth & -          & 48.4 & SGNet \cite{chen2021spatial}*            & ResNet50    & Depth & 76.8       & 51.1                           \\ 
MMAF-Net \cite{fooladgar2019multi} & ResNet101 & Depth & 72.2       & 44.8 & SA-Gate \cite{chen2020bi}*               & ResNet101   & Depth & 77.9       & 52.4                           \\ 
ACNet \cite{hu2019acnet}          & ResNet50  & Depth & -          & 48.3 & InverseForm \cite{borse2021inverseform}* & \textbf{Transformer} & Depth & 78.1       & 53.1                           \\ 
ShapeConv \cite{cao2021shapeconv}  & ResNet101 & HHA   & 75.8       & 50.2 & ShapeConv \cite{cao2021shapeconv}*       & ResNet101   & HHA   & 76.4       & 51.3                           \\ 
ADSD \cite{zhang2022attention}     & ResNet50  & Depth & 77.5       & 52.5 & M2.5D \cite{xing2020malleable}*          & ResNet101   & Depth & 76.9       & 50.9                           \\ 
\rowcolor[HTML]{C0C0C0}Ours                        & ResNet101 & Depth & \textbf{77.7}       & \textbf{52.7} & Ours*                             & ResNet101   & Depth & \textbf{78.4}       & \textbf{53.5} \\ \bottomrule
\end{tabular}}
\label{table1}
\end{table*}

\begin{table*}[]
\centering
\caption{Performance comparison with the state-of-the-art methods on SUN RGB-D datasets.  IS: Information source. `*': multi-scale strategy.}
\resizebox{0.95\linewidth}{!}{
\begin{tabular}{c|c|c|c|c|c|c|c|c|c}
\toprule
Method                                 & Backbone  & IS    & Pixel Acc. & mIoU & Method                        & Backbone  & IS    & Pixel Acc. & mIoU                           \\ \midrule
3DGNN \cite{qi20173d}                         & VGG16     & HHA   & -          & 44.1 & 3DGNN \cite{qi20173d}*               & VGG16     & HHA   & -          & 45.9                           \\ 
D-CNN \cite{wang2018}                    & VGG16     & HHA   & -          & 42.0 & DeepLab-LFOV \cite{chen2017deeplab}* & VGG16     & Depth & 71.9       & 32.1                           \\ 
MMAF-Net \cite{fooladgar2019multi}            & ResNet50  & Depth & 81.0       & 47.0 & RDF \cite{park2017rdfnet}*           & ResNet152 & HHA   & 81.5       & 47.7                           \\ 
SGNet \cite{chen2021spatial}                  & ResNet101 & Depth & 81.0       & 47.5 & SGNet \cite{chen2021spatial}*        & ResNet101 & Depth & 82.0       & 47.6                           \\ 
ShapeConv \cite{cao2021shapeconv}             & ResNet101 & HHA   & 82.0       & 47.6 & ShapeConv \cite{cao2021shapeconv}*   & ResNet101 & HHA   & 82.2       & 48.6                           \\ 
ACNet \cite{hu2019acnet}                      & ResNet50  & HHA   & -          & 48.1 & RefineNet \cite{lin2017refinenet}*   & ResNet152 & Depth & -          & 48.1                           \\ 
ESANet-R34-NBt1D \cite{seichter2021efficient} & ResNet50  & Depth & -          & 48.2 & CRF \cite{lin2017cascaded}*          & ResNet152 & Depth & -          & 48.1                           \\ 
\rowcolor[HTML]{C0C0C0}Ours                                   & ResNet101 & Depth & \textbf{82.4}       & \textbf{49.2} & Ours*                         & ResNet101 & Depth & \textbf{83.3}       & \textbf{49.6} \\ \bottomrule
\end{tabular}}
\label{sunrgbd}
\end{table*}

\begin{table}
    \centering
    \caption{Single-scale testing performance comparison with different baseline methods on NYUDv2 test set. }
    \resizebox{0.95\linewidth}{!}{
        \begin{tabular}{c|c|ccc}
            \toprule
            Method & Backbone                   & Setting  & Pixel Acc. & mIoU \\ \midrule
            \multirow{6}{*}{\makecell{Deeplabv3+                               \\ \cite{chen2018encoder}}} & \multirow{3}{*}{ResNet-101} & baseline & 75.1       & 47.4 \\ \cline{3-5}
                   &                            & ours     & 77.7       & 52.7 \\ \cline{3-5}
                   &                            & $\uparrow (\%)$        & 2.6        & 5.3  \\ \cline{2-5}
                   & \multirow{3}{*}{ResNet-50} & baseline & 74.5       & 46.5 \\ \cline{3-5}
                   &                            & ours     & 76.5       & 51.1 \\ \cline{3-5}
                   &                            & $\uparrow (\%)$         & 2.0        & 4.6  \\ \hline
            \multirow{6}{*}{\makecell{Deeplabv3                                \\ \cite{chen2017rethinking}}}  & \multirow{3}{*}{ResNet-101} & baseline & 73.3       & 45.4 \\ \cline{3-5}
                   &                            & ours     & 75.9       & 50.2 \\ \cline{3-5}
                   &                            & $\uparrow (\%)$         & 2.6        & 4.8  \\ \cline{2-5}
                   & \multirow{3}{*}{ResNet-50} & baseline & 72.7       & 45.2 \\ \cline{3-5}
                   &                            & ours     & 75.6       & 49.3 \\ \cline{3-5}
                   &                            & $\uparrow (\%)$         & 2.9        & 4.1  \\ \hline
            \multirow{6}{*}{\makecell{PSPNet                                   \\ \cite{lin2017feature}}}     & \multirow{3}{*}{ResNet-101} & baseline & 72.8       & 44.3 \\ \cline{3-5}
                   &                            & ours     & 74.9       & 49.1 \\ \cline{3-5}
                   &                            & $\uparrow (\%)$         & 2.1        & 4.8  \\ \cline{2-5}
                   & \multirow{3}{*}{ResNet-50} & baseline & 72.2       & 43.6 \\ \cline{3-5}
                   &                            & ours     & 75.1       & 48.5 \\ \cline{3-5}
                   &                            & $\uparrow (\%)$         & 2.9        & 3.9 \\ \hline
            \multirow{6}{*}{\makecell{FPN                                      \\ \cite{zhao2017pyramid}}}        & \multirow{3}{*}{ResNet-101} & baseline & 74.4       & 46.5 \\ \cline{3-5}
                   &                            & ours     & 76.4       & 50.1 \\ \cline{3-5}
                   &                            & $\uparrow (\%)$         & 2.0        & 4.6  \\ \cline{2-5}
                   & \multirow{3}{*}{ResNet-50} & baseline & 74.1       & 46.2 \\ \cline{3-5}
                   &                            & ours     & 75.9      & 50.1 \\ \cline{3-5}
                   &                            & $\uparrow (\%)$         & 1.8        & 3.9  \\ \bottomrule
        \end{tabular}}
    \label{architectures}
\end{table}


\begin{table}[]
\centering
\caption{The effectiveness of PDC and CPDC on NYUDv2 dataset under single-scale testing. Note: Our proposed method applies PDC for Depth branch and CPDC for RGB branch, and `Place' indicates whether the two modules are placed in the right position.}
\resizebox{0.95\linewidth}{!}{
\begin{tabular}{c|ccccc}
\toprule
Method & Depth branch & RGB branch  & Place  & Pixel Acc.                     & mIoU                           \\ \midrule
$\mathfrak{a}$\ & Baseline     & Baseline   &-   & 75.1                           & 47.4                           \\
$\mathfrak{b}$ & Baseline + PDC & Baseline   &yes   & 76.9                           & 51.8                          \\
$\mathfrak{c}$ & Baseline     & Baseline + CPDC  &yes & 76.8                           & 51.6                           \\
$\mathfrak{d}$ & Baseline + CPDC & Baseline  & no  & 75.7 & 50.7 \\

$\mathfrak{e}$  & Baseline      & Baseline + PDC &no & 76.2       & 50.9 \\
$\mathfrak{f}$ (Ours)   & Baseline + PDC & Baseline + CPDC &yes & \cellcolor[HTML]{EFEFEF}\textbf{77.7} & \cellcolor[HTML]{EFEFEF}\textbf{52.7} \\ \bottomrule
\end{tabular}}
\label{Ablation1}
\end{table}



\begin{table}[]
\caption{Comparison with other context aggregation methods on NYUDv2 dataset. The four methods are for RGB data to capture long-range dependence and no operations are performed on the Depth data.}
\resizebox{0.95\linewidth}{!}{
\begin{tabular}{c|cccccc}
\toprule
Method  & PLK & NL & CLK & CPDC (CLK+PDC)  & Pixel Acc. & mIoU \\ \midrule
$\mathfrak{a}$  &  \Checkmark         &     &     &     & 75.1      & 47.8 \\
$\mathfrak{b}$      &          & \Checkmark    &   &       & 75.5       & 49.5 \\
$\mathfrak{c}$      &          &     & \Checkmark    &     & 75.8       & 50.3 \\ 
$\mathfrak{d}$      &          &     &   &  \Checkmark     &\cellcolor[HTML]{EFEFEF} 76.8      & \cellcolor[HTML]{EFEFEF} 51.6 \\ 
  \bottomrule
\end{tabular}}
\label{Ablation3}
\end{table}

\section{Experiments}
\label{Experiment}

\subsection{Dataset and Metrics}

\textbf{NYUDv2} \cite{silberman2012indoor} consists of 1,449 densely labeled pairs of RGB-D images.
We keep the 40-class settings, where 795 images are used for training and the remaining 654 images are used for testing.

\textbf{SUN RGB-D} \cite{song2015sun} contains 10,335 RGB-D images with 37 categories. We follow the common setting, where 5,285 images are used as training set and 5,050 are utilized as testing set.

We apply two common metrics, $i.e.,$ Pixel Accuracy (Pixel Acc.), and Mean Intersection Over Union (mIoU) to report the results.
The two metrics are defined as follows: 
\begin{equation}
\text{Pixel Acc.}=\sum_i\frac{p_{ii}}{g} ,    
\end{equation}
\begin{equation}
\text{mIoU}=\frac{1}{M}\sum_i\frac{p_{ii}}{g_i+\sum_{j}p_{ji}-p_{ii}}.     
\end{equation}
where $p_{ij}$ is the number of pixels that are predicted as class $j$ with ground truth class $i$, $M$ is the number of classes and $g_i$ is the number of pixels with ground truth class $i$, the total number of all pixels is $g=\sum_{i}g_i$.

\begin{figure*}[t]
    \centering
    \begin{minipage}[c]{0.158\linewidth}
        \includegraphics[width=1.15in]{pic/visualize/nyuv2/189_rgb.jpg}
    \end{minipage}
    \begin{minipage}[c]{0.158\linewidth}
        \includegraphics[width=1.15in]{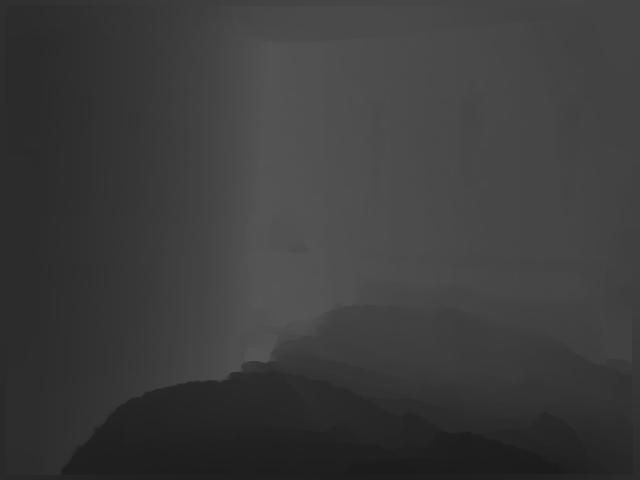}
    \end{minipage} 
    \begin{minipage}[c]{0.158\linewidth}
        \includegraphics[width=1.15in]{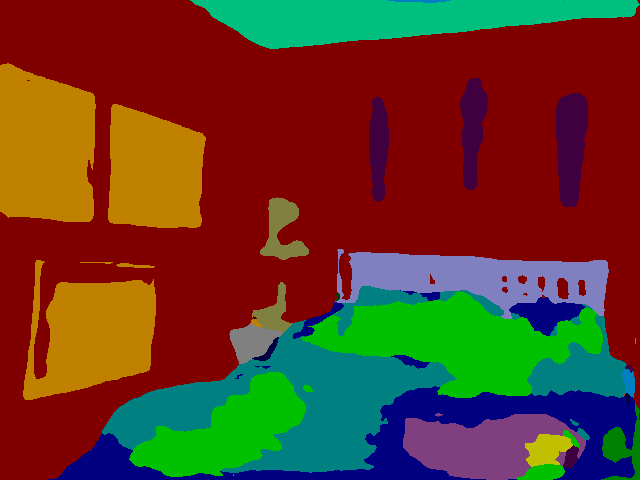}
    \end{minipage}
    \begin{minipage}[c]{0.158\linewidth}
        \includegraphics[width=1.15in]{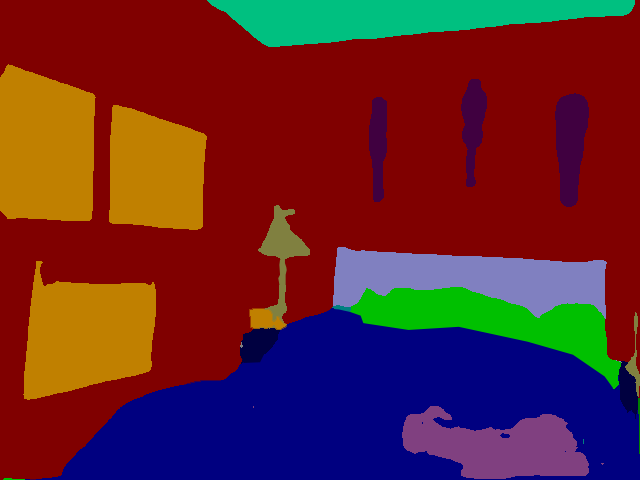}
    \end{minipage} 
    \begin{minipage}[c]{0.158\linewidth}
        \includegraphics[width=1.15in]{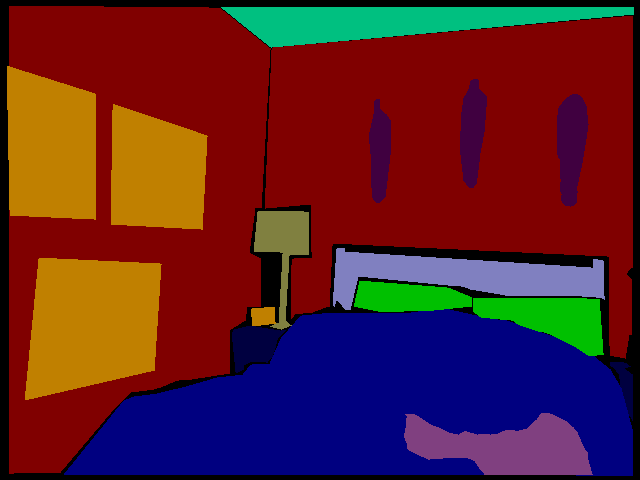}
    \end{minipage} 
    
    \begin{minipage}[c]{0.158\linewidth}
        \includegraphics[width=1.15in]{pic/visualize/nyuv222/411_rgb.jpg}
    \end{minipage} 
    \begin{minipage}[c]{0.158\linewidth}
        \includegraphics[width=1.15in]{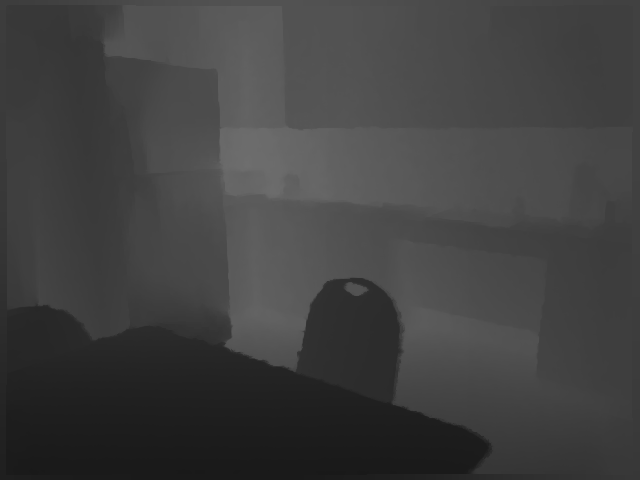}
    \end{minipage} 
    \begin{minipage}[c]{0.158\linewidth}
        \includegraphics[width=1.15in]{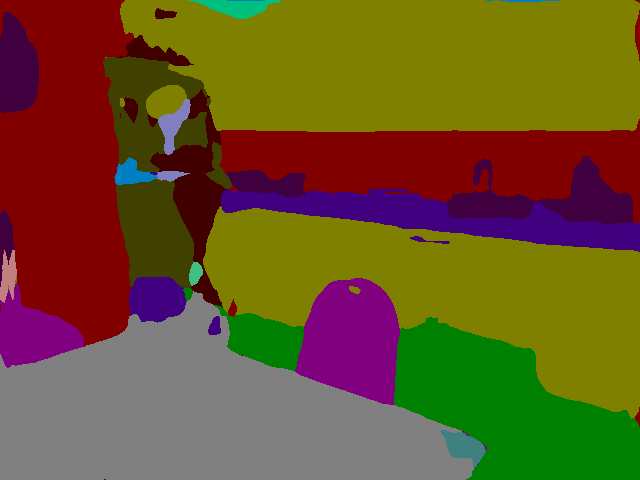}
    \end{minipage} 
    \begin{minipage}[c]{0.158\linewidth}
        \includegraphics[width=1.15in]{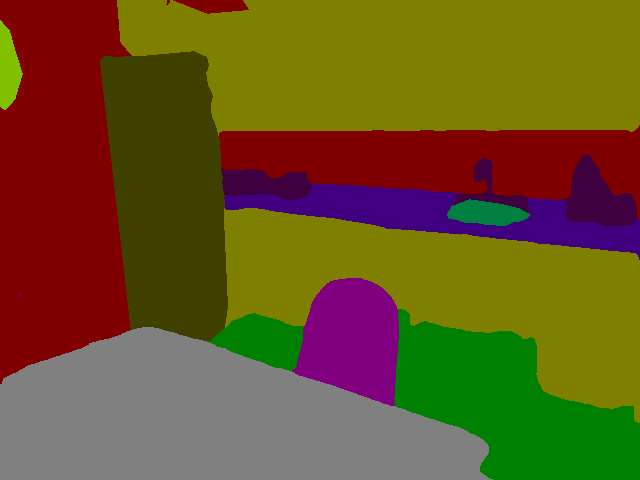}
    \end{minipage} 
    \begin{minipage}[c]{0.158\linewidth}
        \includegraphics[width=1.15in]{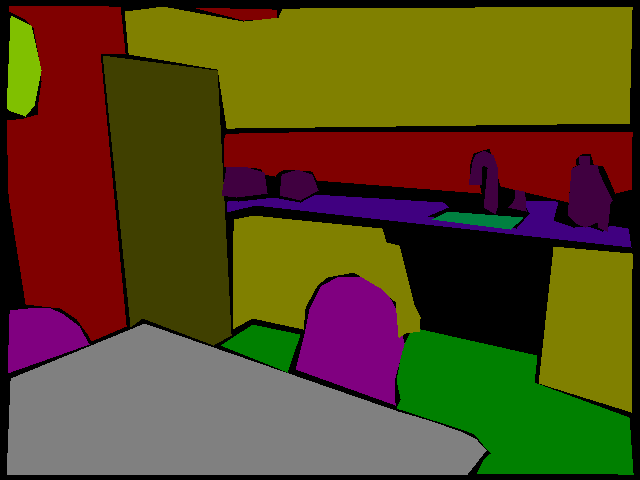}
    \end{minipage}
    
    \begin{minipage}[c]{0.158\linewidth}
        \includegraphics[width=1.15in]{pic/visualize/nyuv222/1074_rgb.jpg}
    \end{minipage} 
    \begin{minipage}[c]{0.158\linewidth}
        \includegraphics[width=1.15in]{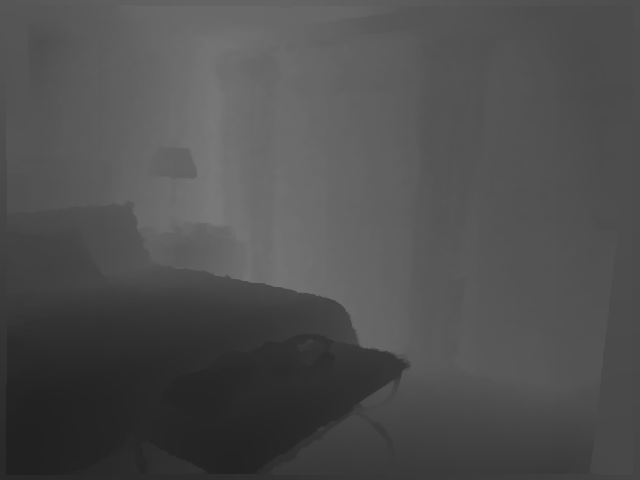}
    \end{minipage} 
    \begin{minipage}[c]{0.158\linewidth}
        \includegraphics[width=1.15in]{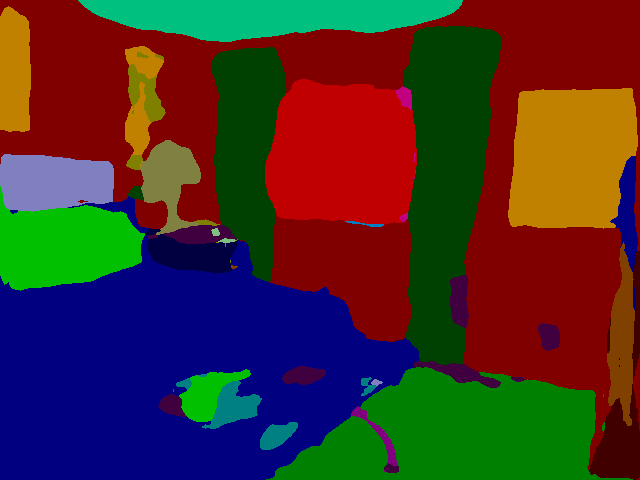}
    \end{minipage} 
    \begin{minipage}[c]{0.158\linewidth}
        \includegraphics[width=1.15in]{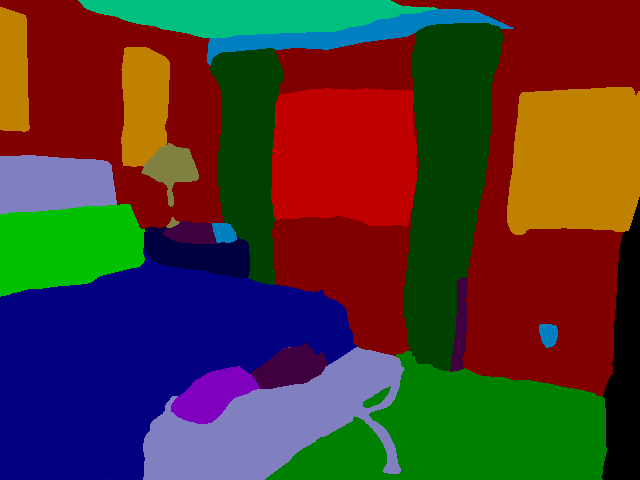}
    \end{minipage}
    \begin{minipage}[c]{0.158\linewidth}
        \includegraphics[width=1.15in]{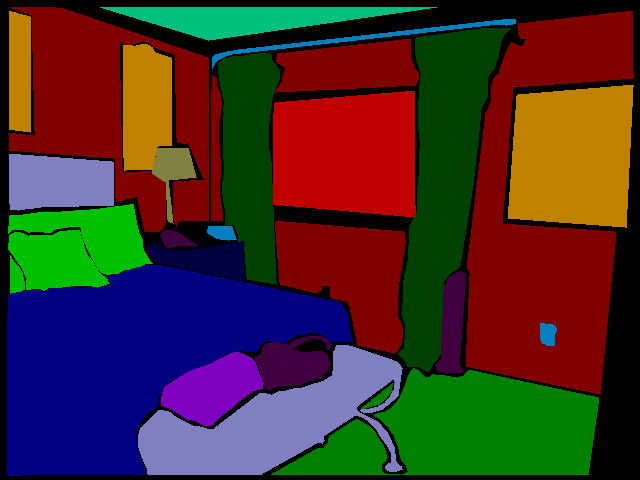}
    \end{minipage}
    
    \begin{minipage}[c]{0.158\linewidth}
        \includegraphics[width=1.15in]{pic/visualize/nyuv222/1390_rgb.jpg}
    \centerline{RGB}
    \end{minipage} 
    \begin{minipage}[c]{0.158\linewidth}
        \includegraphics[width=1.15in]{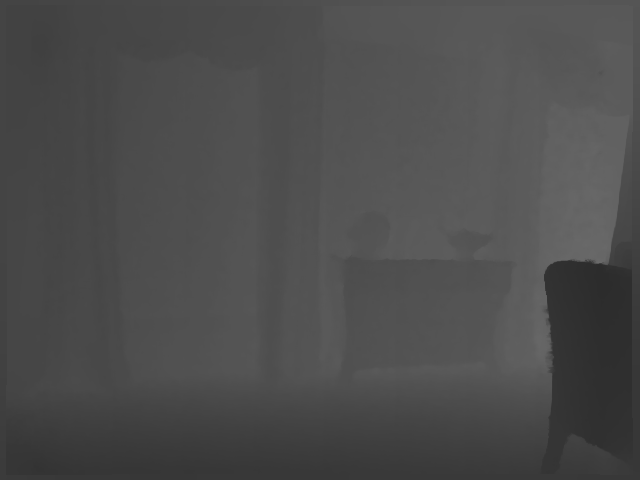}
    \centerline{Depth}
    \end{minipage}
    \begin{minipage}[c]{0.158\linewidth}
        \includegraphics[width=1.15in]{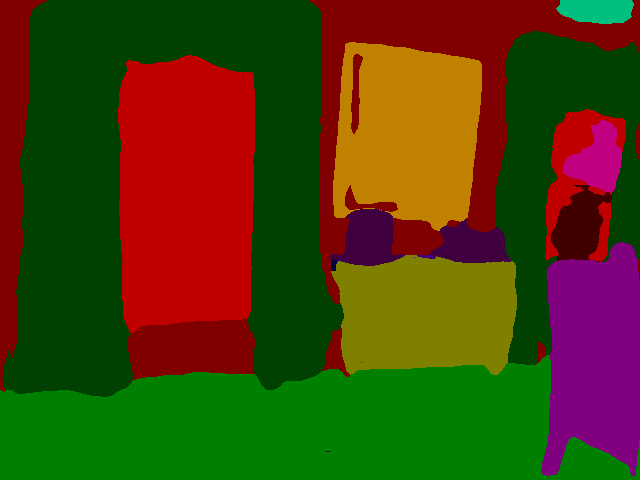}
    \centerline{Baseline}
    \end{minipage}
    \begin{minipage}[c]{0.158\linewidth}
        \includegraphics[width=1.15in]{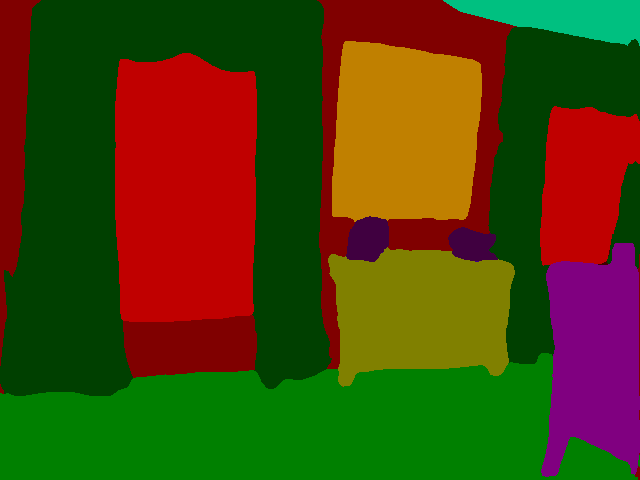}
    \centerline{Ours}
    \end{minipage} 
    \begin{minipage}[c]{0.158\linewidth}
        \includegraphics[width=1.15in]{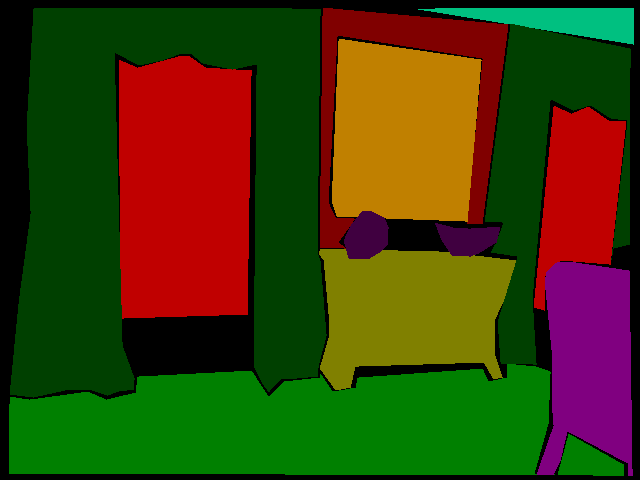}
    \centerline{Ground truth}
    \end{minipage}
    \caption{Visual comparison of scene semantic segmentation on NYUDv2 dataset.}
    \label{NYUDv2_result}
\end{figure*}

\begin{figure*}[t]
    \centering
    \begin{minipage}[c]{0.158\linewidth}
        \includegraphics[width=1.15in]{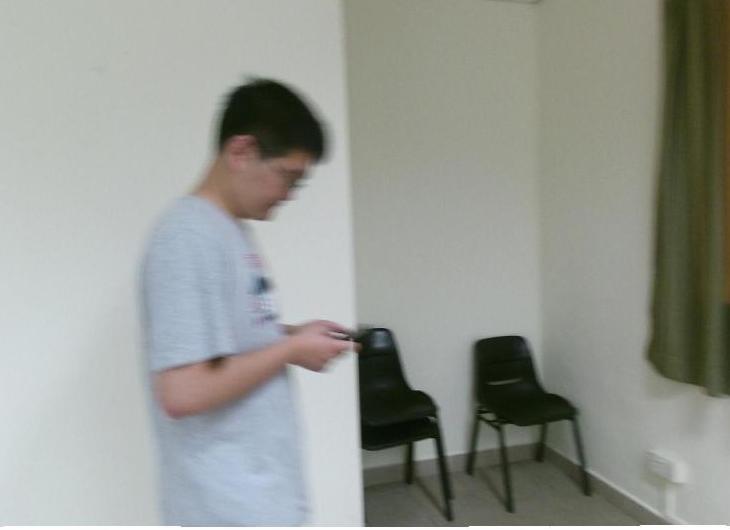}
    \end{minipage}
    \begin{minipage}[c]{0.158\linewidth}
        \includegraphics[width=1.15in]{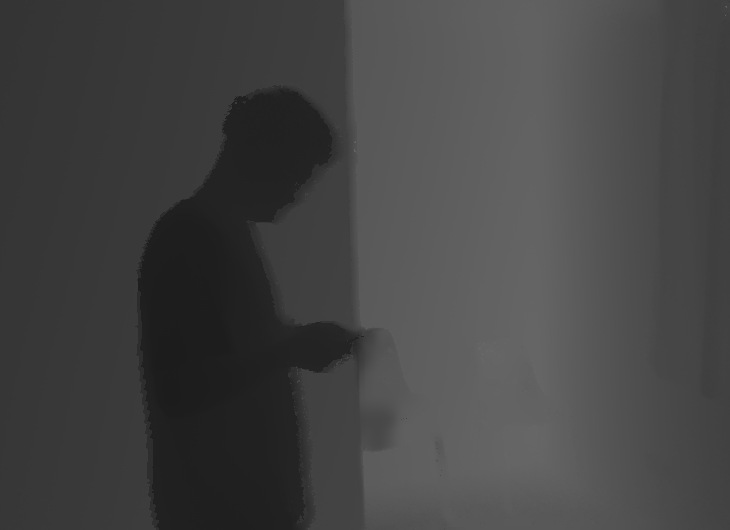}
    \end{minipage} 
    \begin{minipage}[c]{0.158\linewidth}
        \includegraphics[width=1.15in]{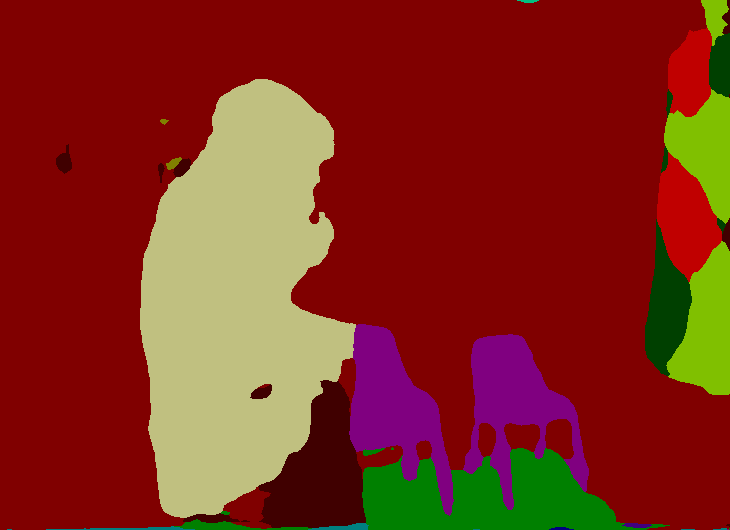}
    \end{minipage} 
    \begin{minipage}[c]{0.158\linewidth}
        \includegraphics[width=1.15in]{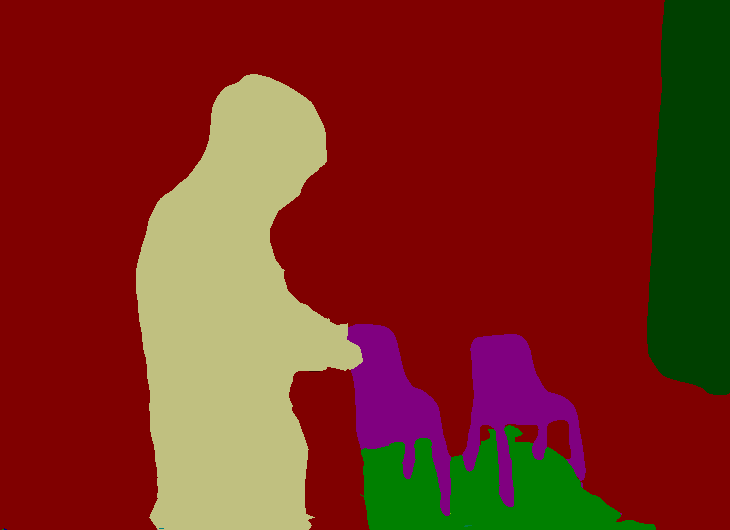}
    \end{minipage} 
    \begin{minipage}[c]{0.158\linewidth}
        \includegraphics[width=1.15in]{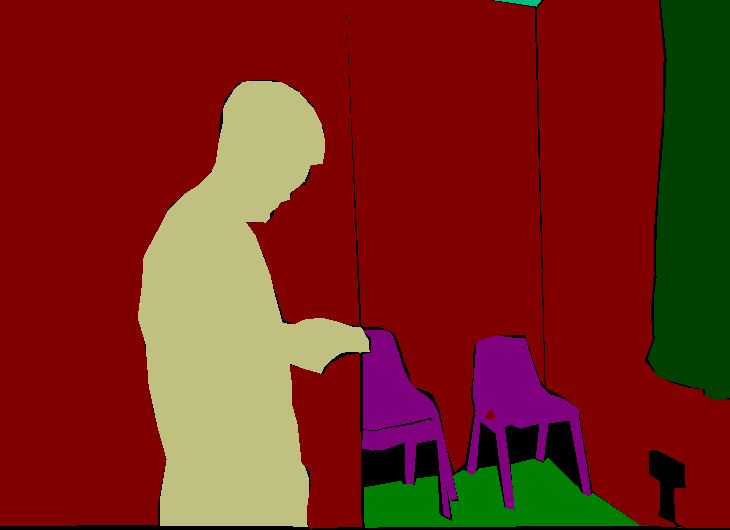}
    \end{minipage} 
    
    \begin{minipage}[c]{0.158\linewidth}
        \includegraphics[width=1.15in]{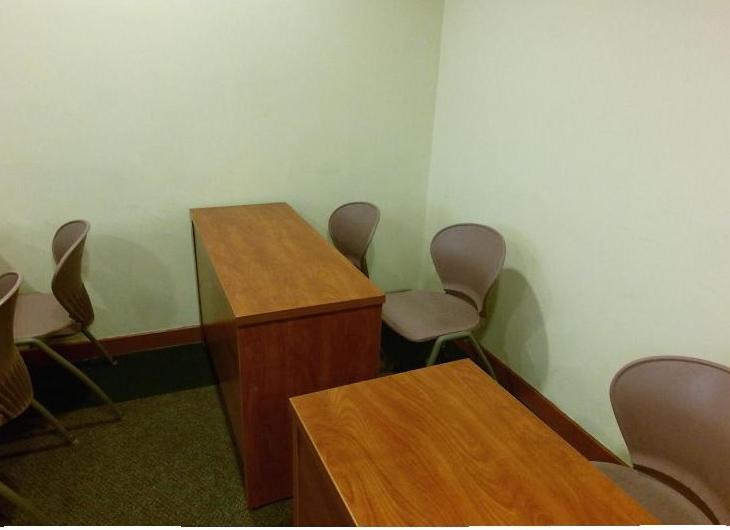}
    \end{minipage}
    \begin{minipage}[c]{0.158\linewidth}
        \includegraphics[width=1.15in]{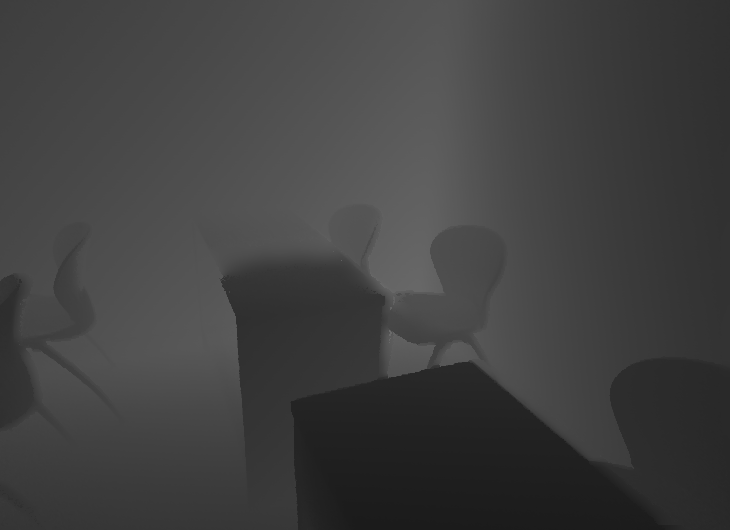}
    \end{minipage} 
    \begin{minipage}[c]{0.158\linewidth}
        \includegraphics[width=1.15in]{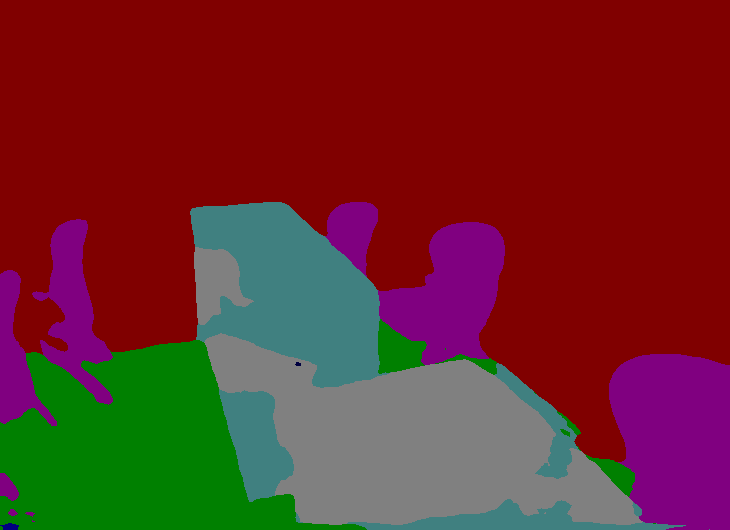}
    \end{minipage} 
    \begin{minipage}[c]{0.158\linewidth}
        \includegraphics[width=1.15in]{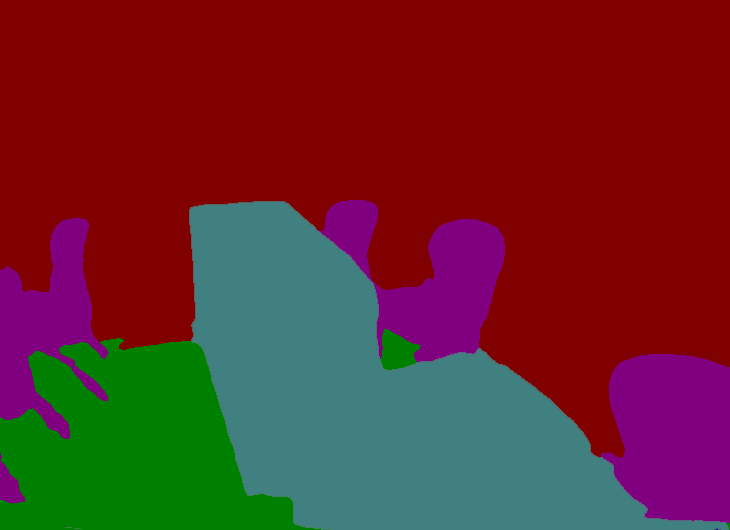}
    \end{minipage} 
    \begin{minipage}[c]{0.158\linewidth}
        \includegraphics[width=1.15in]{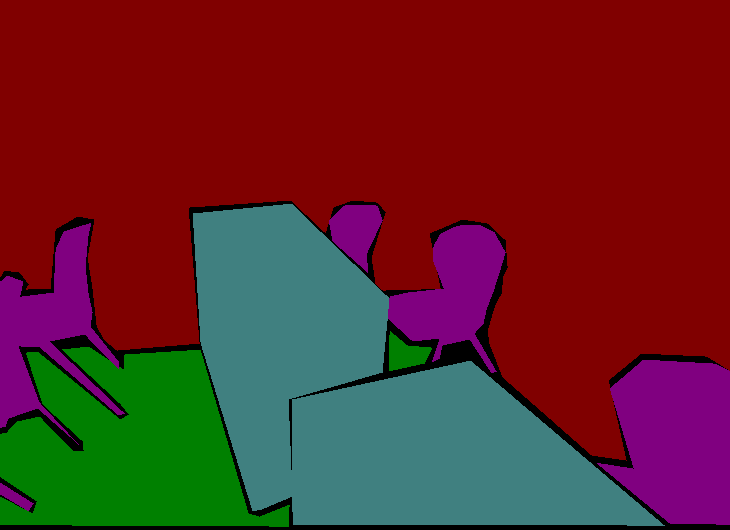}
    \end{minipage}
    
    \begin{minipage}[c]{0.158\linewidth}
        \includegraphics[width=1.15in]{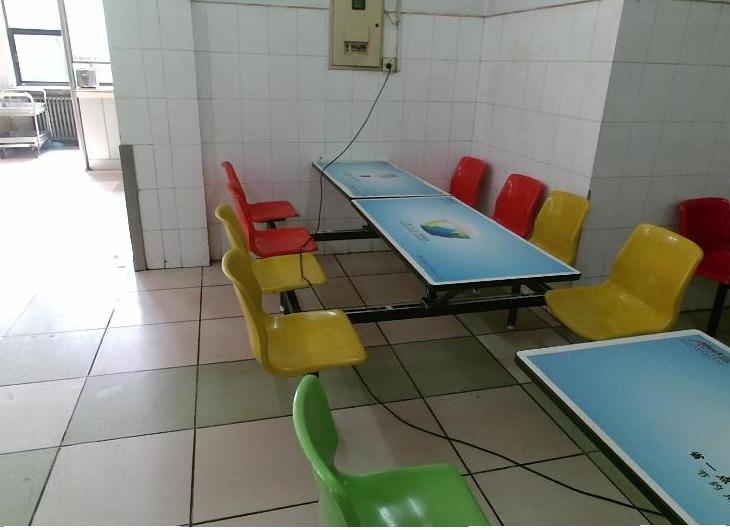}
    \end{minipage} 
    \begin{minipage}[c]{0.158\linewidth}
        \includegraphics[width=1.15in]{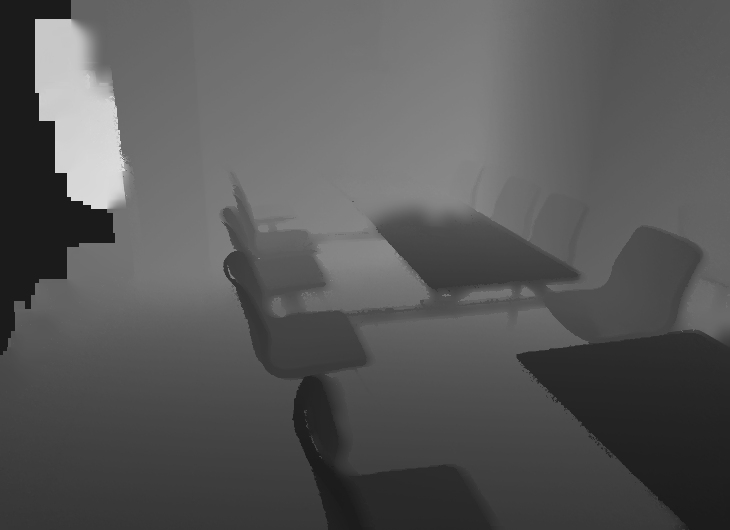}
    \end{minipage}
    \begin{minipage}[c]{0.158\linewidth}
        \includegraphics[width=1.15in]{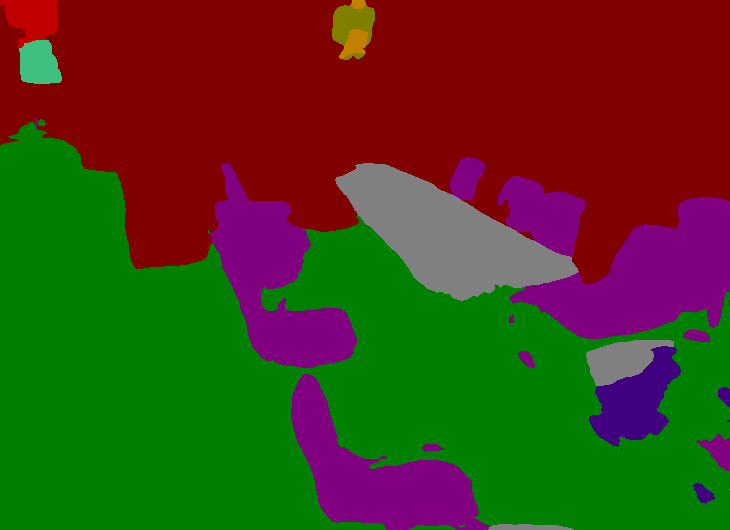}
    \end{minipage} 
    \begin{minipage}[c]{0.158\linewidth}
        \includegraphics[width=1.15in]{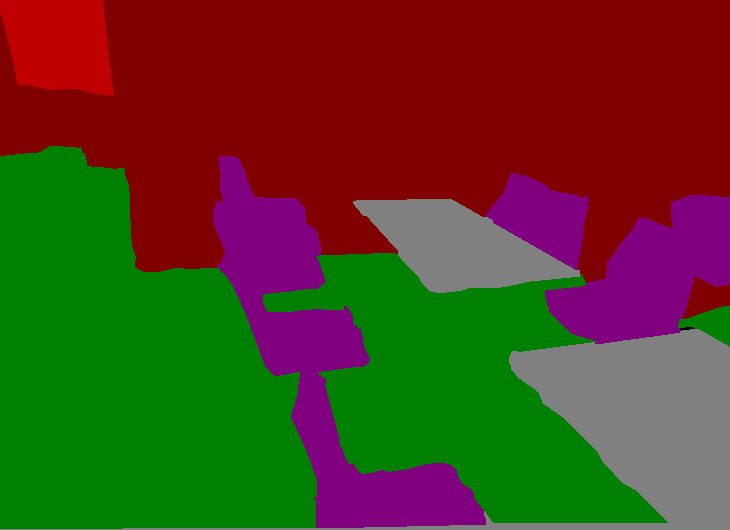}
    \end{minipage} 
    \begin{minipage}[c]{0.158\linewidth}
        \includegraphics[width=1.15in]{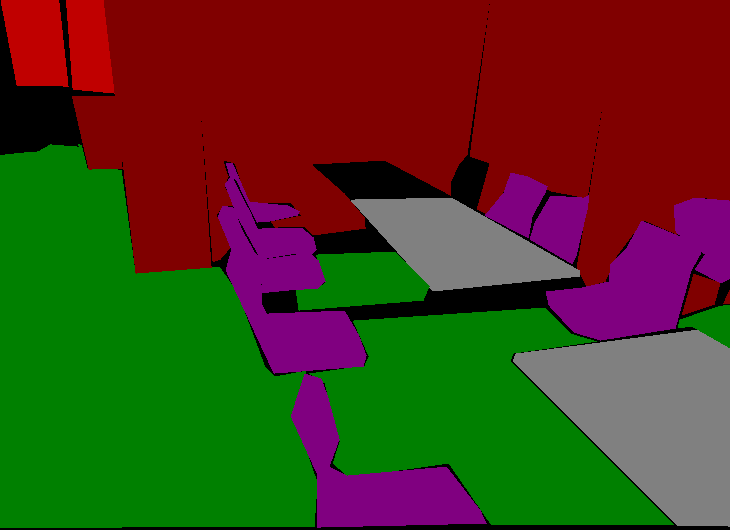}
    \end{minipage}
    
    \begin{minipage}[c]{0.158\linewidth}
        \includegraphics[width=1.15in]{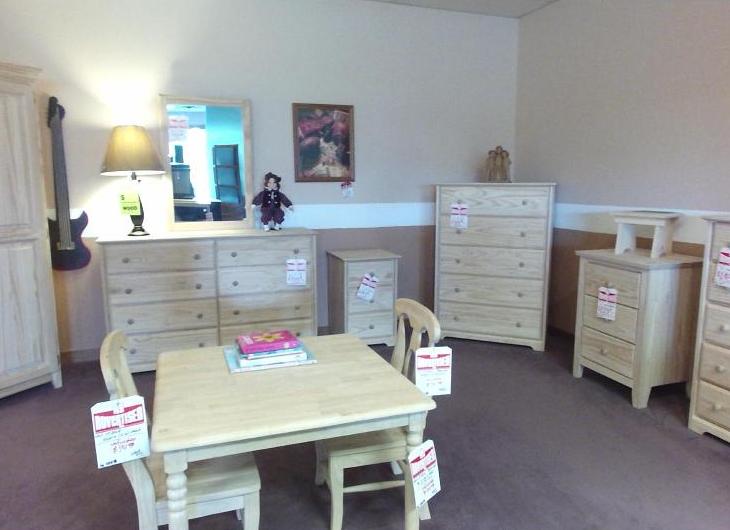}
    \centerline{RGB}
    \end{minipage} 
    \begin{minipage}[c]{0.158\linewidth}
        \includegraphics[width=1.15in]{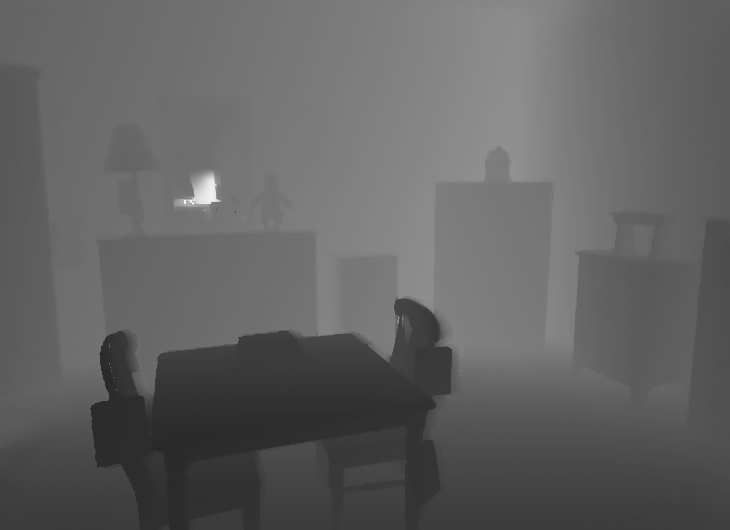}
    \centerline{Depth}
    \end{minipage} 
    \begin{minipage}[c]{0.158\linewidth}
        \includegraphics[width=1.15in]{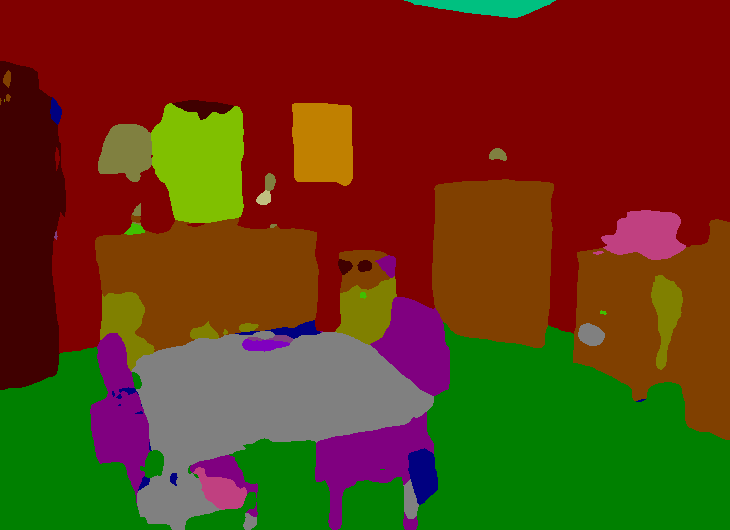}
    \centerline{Baseline}
    \end{minipage} 
    \begin{minipage}[c]{0.158\linewidth}
        \includegraphics[width=1.15in]{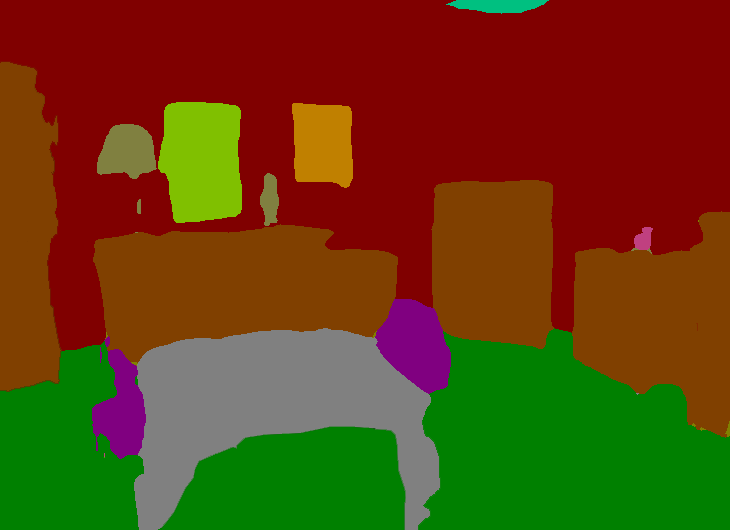}
    \centerline{Ours}
    \end{minipage} 
    \begin{minipage}[c]{0.158\linewidth}
        \includegraphics[width=1.15in]{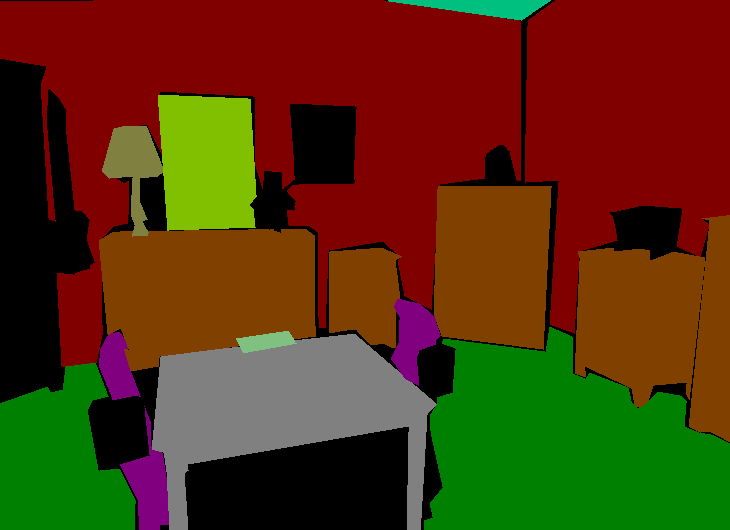}
    \centerline{Ground truth}
    \end{minipage}
    \caption{Visual comparison of scene semantic segmentation on SUN RGB-D dataset.}
    \label{SUN RGB-D_result}
\end{figure*}

\begin{figure*}
    \centering
    \begin{minipage}[c]{1\linewidth}
        \includegraphics[scale=0.69]{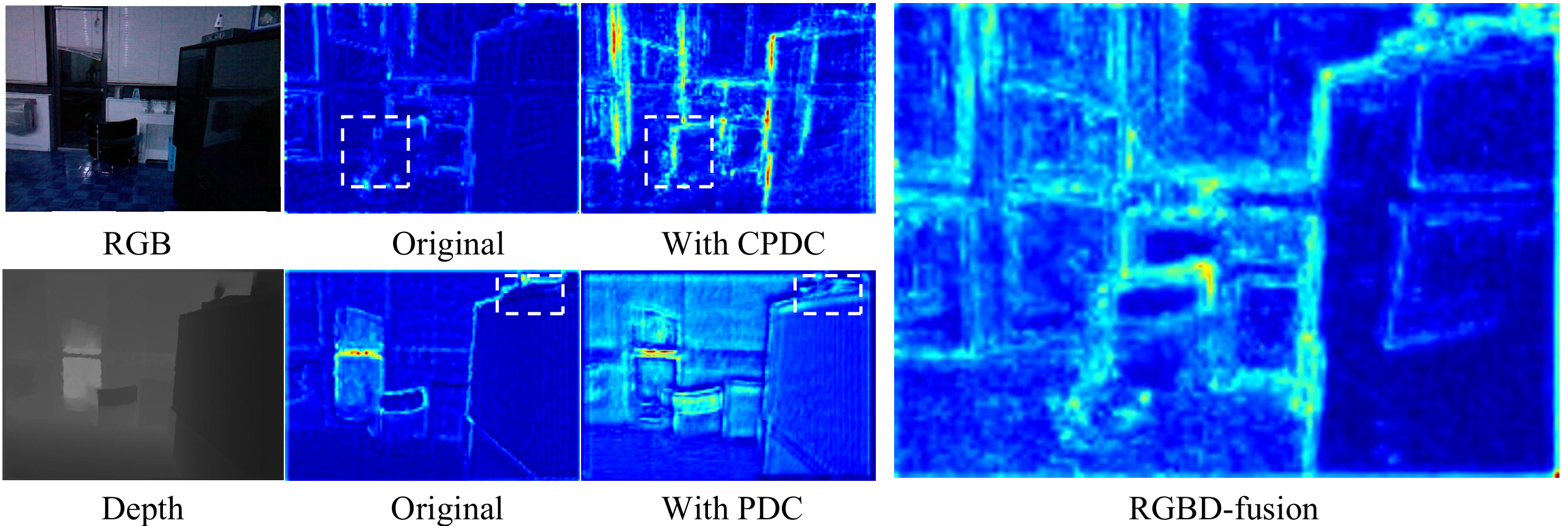}
    \end{minipage}
    \begin{minipage}[c]{1\linewidth}
        \includegraphics[scale=0.69]{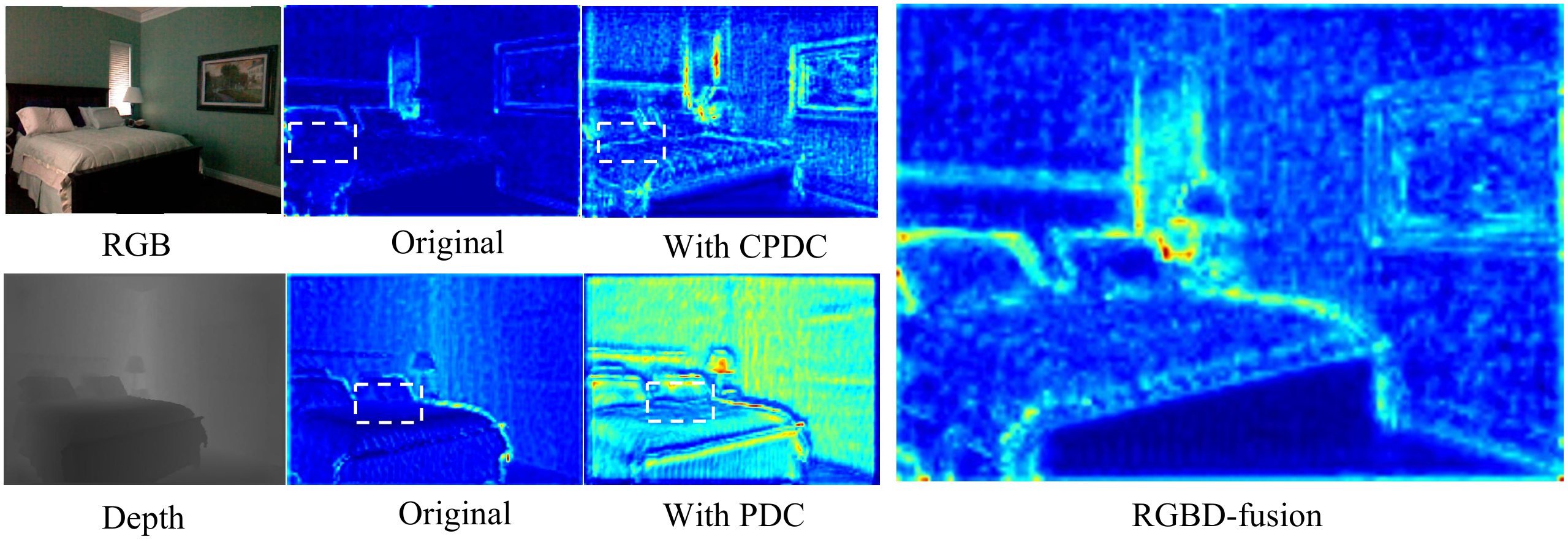}
    \end{minipage}
        \begin{minipage}[c]{1\linewidth}
        \includegraphics[scale=0.69]{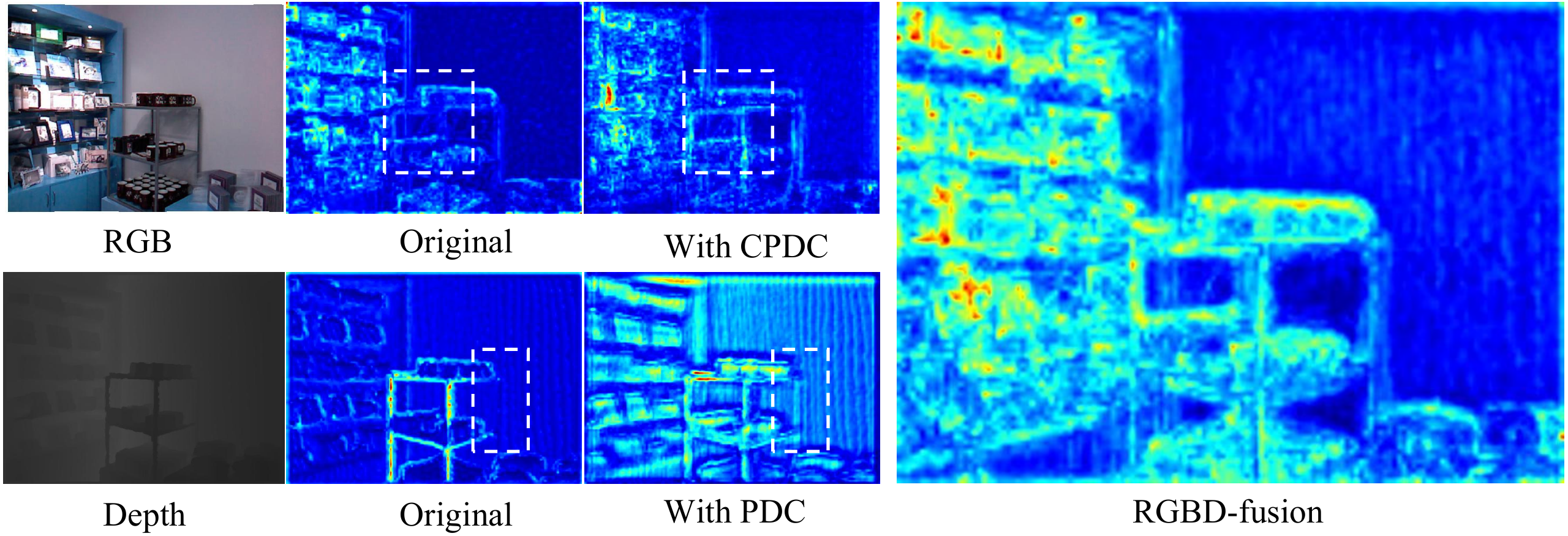}
    \end{minipage}
    \caption{Visualizations of feature map response. We compare the response map of Depth data with/without PDC and RGB data with/without CPDC.}
    \label{feature_map}
\end{figure*}

\subsection{Implementation Details}
\subsubsection{Training Setting} We implement our network using the PyTorch deep learning framework \cite{paszke2019pytorch}, and train the models on two NVIDIA Tesla V100 GPUs. The initial learning rate is set to 8e-3 with the ``poly” policy \cite{liu2015parsenet}. The input images are cropped to $480 \times 480$ and the batch size is $8$. The SGD is used as our optimizer, with momentum of 0.9 and a weight decay of 1e-4.

\subsubsection{Comparison Protocol} For fair comparisons, we compare the baseline (two original parallel DeepLabv3+, ResNet-101) \cite{chen2018encoder} with two improved encoders (equipped with the PDC and CPDC, respectively), without changing other settings in DeepLabv3+. This is to verify that the model effectiveness comes from the proposed PDC and CPDC, not other factors.

For a fair comparison with other state-of-the-arts, we employ both single-scale and multi-scale testing strategies. If not otherwise noted, the experiments are evaluated under single-scale testing, and `$*$' in tables indicates the multi-scale strategy.

\subsection{Comparisons with State-of-the-arts}
\subsubsection{NYUDv2} As shown in Tab.~\ref{table1}, we achieve the Pixel Acc and mIoU of 77.7\% and 52.7\% under single-scale testing, outperforming the contemporary work, ShapeConv \cite{cao2021shapeconv} by 1.9\% in and 2.5\% in Pixel Acc. and mIoU, respectively. Moreover, our segmentation performance is 0.3\% Pixel Acc. and 0.4\%
mIoU more than InverseForm \cite{borse2021inverseform} using Transformer as the backbone. On multi-scale testing, our PDCNet also achieves leading performance.

\subsubsection{SUN RGB-D} Tab.~\ref{sunrgbd} shows the testing results on SUN RGB-D dataset. Our PDCNet outperforms other methods under both single-scale and multi-scale testing. For example, under multi-scale testing, PDCNet outperforms the contemporary work, SGNet \cite{chen2021spatial} by 1.3\% in Pixel Acc and 2.0\% in mIoU, respectively.
The competitive performances benefit from our proposed PDC and CPDC, which can make use of subtle pixel-wise gradient-level and intensity-level information in local and global range simultaneously. In other words, the PDC and CPDC exploit the characteristics of the two modal data and complement each other effectively.

\subsection{Experiments on Different Architectures}
To confirm the generalization performance of our method, we equip the PDC and CPDC modules with other classic semantic segmentation architectures, including Deeplabv3+ \cite{chen2018encoder}, Deeplabv3 \cite{chen2017rethinking}, PSPNet \cite{lin2017feature}
and FPN \cite{zhao2017pyramid} with (ResNet-50, ResNet-101 \cite{he2016deep}) on NYUDv2 dataset. As shown in Tab.~\ref{architectures}, under all settings, our method achieves signiﬁcant performance improvements, demonstrating its generalization ability. For example, when applying Deeplabv3+ as the baseline, the
results reveal that our model achieves significant improvements in performance (+2.6\%
Pixel Acc and +5.3\% mIoU). 

\subsection{Ablation Study}
\label{Ablation Study}
\subsubsection{Effectiveness of PDC and CPDC}
Tab.~\ref{Ablation1} shows the indispensability of our model on the NYUDv2 dataset by adopting the proposed PDC and CPDC modules. As shown in Tab.~\ref{Ablation1}, we can conclude that 1) comparing our model with $\mathfrak{a}$ to $\mathfrak{c}$, both the PDC and CPDC modules improve model performance and work best when used in combination. 2) Comparing our model with $\mathfrak{d}$ and $\mathfrak{e}$ illustrate that our original setting, $i.e.,$ PDC and CPDC are suitable for Depth and RGB data, respectively. This also explains that Depth maps are more suitable for providing intrinsic fine-grained patterns of objects due to their Depth continuity, while RGB images effectively provide a global view.

In this regard, the PDC and CPDC are complementary and can function as a unified unit to mine the intra-domain inherent context within the data. The model is enabled to explore the inherent intra-domain context by explicitly regularizing the feature space via aggregation of gradient pixel-wise differences and intensity pixel-wise differences.


\subsubsection{Comparison with Alternatives} 
One may argue that our self-attention based modules can replace our CLK to capture long-range contextual information. Here, we compare the performance of our CLK with a classic self-attention module, Non-local neural networks \cite{wang2018non}, denoted as `NL'. We compare the long-range context aggregation abilities in the RGB branch, while Depth branch utilizes original baseline backbone. As can be observed in Tab.~\ref{Ablation3} $\mathfrak{b}$ and $\mathfrak{c}$, CLK outperforms `NL' by 0.3\% Pixel Acc. and 0.8\% mIoU. Combing CLK and PDC, $i.e.,$ CPDC, achieves the best results.

In fact, self-attention ignores the fundamental difference between sequence-based NLP tasks and the 2D structure, which destroys the crucial 2D structure of images. Moreover, it is worth noting that channels-wise information often represents different objects in CNNs \cite{qin2020ffa,chen2017sca, ma2022learning} and the channel-wise adaptability is also crucial for segmentation tasks, which is lacking in self-attention. With depth-wise convolution, each input channel undergoes a convolution separately, while point-wise convolution is performed to combine the outputs from depth-wise convolution.
Compared with `NL', CLK not only involves the advantages of long-range dependence in 2D structure, but also achieves channel-wise adaptability.

We also verify the performance of our CLK by comparing it with the Parallel mode large kernel shown in Eq.~\ref{mode}, denoted as `PLK'. As illustrated in $\mathfrak{a}$ and $\mathfrak{c}$ of Tab.~\ref{Ablation3}, compared with `PLK', the CLK has a larger receptive field without losing any local information and henceforce achieves higher Pixel Acc and mIoU. Notably, the dilation of $7 \times 7$ convolution can be filled by the $5 \times 5$ convolution, which means the CLK do not lacks any information.

\subsubsection{Impact of $\alpha$ in PDCNet} As discussed in Eq.(\ref{pdc}) that $\alpha$ adjusts the contribution of fine-grained gradient-level changes and intensity-level changes for semantic information. As illustrated in Tab.~\ref{impact}, when $\alpha$ is set as a learnable parameter, both Pixel Acc. and mIoU achieves the best score. Additionally, we also set
the $\alpha$ as default configurations, $i.e.$, varying from 0 to 1 to test the importance of the pixel difference term in PDC. It proves that pixel difference term shown in Eq.~\ref{pdc} is more crucial than the vanilla term, which means integrating pixel difference operation into CNNs is an effective solution. 



\begin{table}
\centering
\caption{Impact of $\alpha$ in PDCNet tested on NYUDv2 dataset under multi-scale testing.}
\label{nyud2}
\resizebox{0.6\linewidth}{!}{
\begin{tabular}{ccccc} 
\toprule

{$\alpha$} & Pixel Acc.    & mIoU           \\  \midrule
0.1 & 77.2       & 51.3   \\
0.2 & 77.4       & 51.5  \\
0.3 & 77.6       & 52.1   \\
0.4 & 77.8       & 52.5  \\
0.5 & 78.0       & 52.8  \\
0.6 & 78.1       & 53.0   \\
0.7 & 78.1       & 53.2   \\
0.8 & \cellcolor[HTML]{EFEFEF}\textbf{78.2}       & \cellcolor[HTML]{EFEFEF}\textbf{53.4}  \\
0.9 & 78.0       & 52.8 \\
\midrule
learnable   & \cellcolor[HTML]{C0C0C0}\textbf{78.4}      & \cellcolor[HTML]{C0C0C0}\textbf{53.5} \\
\bottomrule
\label{impact}
\end{tabular}}
\end{table}

\begin{table}
\centering
\caption{Impact of kernel size in PDC tested on NYUDv2 dataset under multi-scale testing. }
\resizebox{0.6\linewidth}{!}{
\begin{tabular}{ccc} 
\toprule
\multicolumn{1}{l}{Kernel size} & Pixel Acc.    & mIoU           \\ \midrule
$3 \times3$                     & 78.2          & 53.1           \\
$5 \times5$                     & \cellcolor[HTML]{C0C0C0}\textbf{78.4} & \cellcolor[HTML]{C0C0C0}\textbf{53.5}  \\
$7 \times7$                     & 78.4          & 53.0           \\
$9 \times9$                     & 78.1          & 52.9           \\
\bottomrule
\label{kernel_size}
\end{tabular}}
\end{table}

\subsubsection{Impact of Kernel Size in PDC}
Compared with the long-range receptive field in CPDC, our PDC applies the convolution kernel of $5 \times 5$, dilation 1 to capture fine-grained gradient-level and intensity-level semantic information in local regions from Depth data. We conduct the ablation study to compare the effectiveness of convolution kernels of different sizes. As shown in Tab.\ref{kernel_size}, PDC achieves the best performance when applying the kernel size of $5 \times5$. 

\section{Visualization of PDCNet}
To illustrate the results of the model more intuitively, we display the qualitative results of the NYUDv2 and SUN RGB-D datasets in Fig.~\ref{NYUDv2_result} and Fig.~\ref{SUN RGB-D_result}, respectively. From the results, we confirm that the local pixel difference in Depth image and global dependence are well enhanced by our PDCNet. For example, in the first row of Fig.~\ref{NYUDv2_result}, the pillows have very similar patterns to the bed, which cannot be easily discriminated by just the 2D appearance, while they can be well separated by the local pixel difference in Depth. Furthermore, the global dependence makes sure that the rest of the bed is not misidentified as a pillow. Indeed, this is what we find: since CPDC captures features in high contrast areas from RGB images while PDC separates objects by Depth rather than lighting conditions. 

We also visualize the feature maps from the RGB branch and Depth branch of our proposed model as well as the feature maps of the original Deeplabv3+ model. As illustrated in Fig.~\ref{feature_map}, the PDC and CPDC focus more on pixel-level subtle information, such as edges and boundaries. Moreover, the PDC tends to show the overall outline of the object better. Notably, we believe that our PDC overcomes image noise, which can be analyzed from the principle of PDC: If a data point in Depth map is noisy, the pixel differences between this point and other points are very large, which can be easily captured. This is expected, it is also the pixel difference convolution matters.

\section{Conclusion}
We present a state-of-the-art two-branch PDCNet by introducing two plug-and-play modules: PDC and CPDC. PDC considers subtle geometric information (pixel difference) that occurs in local regions in Depth data. We also propose CLK, with which PDC is easily extended to CPDC to absorb the advantage of capturing local pixel-wise features of PDC and long-range dependence of CLK, which further captures refined pixel discrepancy in the satisfying receptive field. Consequently, both intensity and gradient information from 2D patterns and 3D geometric maps are seamlessly incorporated into our PDCNet. We also conduct extensive experiments on NYUDv2 and SUN RGB-D to evaluate the
superiority of our proposed model. Empirical results illustrate that PDCNet exhibits substantial improvements over the state-of-the-arts.

\section*{Acknowledgement} 
This work was supported by the grants from the National Key R\&D Program of China (2019YFC1906201), the National Natural Science Foundation of China (91748122). Jun Yang thanks for China Scholarship Council (CSC) for supporting him to study at École normale supérieure and also thanks for Yerkebulan Massalim, a PH.D student in École normale supérieure, for his help in English improvement.

\bibliographystyle{IEEEtran}




\end{document}